\documentclass{article}
\usepackage{arxiv}
\usepackage[utf8]{inputenc} 
\usepackage[T1]{fontenc}    
\usepackage{hyperref}       
\usepackage{url}            
\usepackage{booktabs}   
\usepackage{caption}
\usepackage[flushleft]{threeparttable}
\usepackage{subcaption} 
\usepackage{amsfonts}       
\usepackage{amsmath}
\usepackage{bbm}
\usepackage{amsthm}
\usepackage{mathtools}
\usepackage{nicefrac}       
\usepackage{microtype}      
\usepackage{lipsum}
\usepackage{algorithm}
\usepackage{algpseudocode}
\usepackage{graphicx} 
\usepackage{xcolor}

\graphicspath{{figures/}}

\newtheorem{prop}{Proposition}
\newcommand{\bs}{\boldsymbol}
\newcommand{\xib}{\bs{\xi}}
\DeclareMathOperator*{\argmax}{argmax}
\DeclareMathOperator*{\argmin}{argmin}
\DeclareMathOperator{\Tr}{Tr}
\newcommand{\norm}[1]{\left\lVert#1\right\rVert}

\title{BayesFlow: Learning complex stochastic models with invertible neural networks}

\author{
  Stefan T.~Radev \\
  Institute of Psychology\\
  Heidelberg University\\
  Hauptstr. 47-51, 69117 Heidelberg \\
  \texttt{stefan.radev93@gmail.com}
   \And
 Ulf K.~Mertens \\
  Institute of Psychology\\
  Heidelberg University\\
  Hauptstr. 47-51, 69117 Heidelberg \\
  \texttt{mertens.ulf@gmail.com}
  \And
 Andreas Voss \\
  Institute of Psychology\\
  Heidelberg University\\
  Hauptstr. 47-51, 69117 Heidelberg \\
  \texttt{andreas.voss@psychologie.uni-heidelberg.de} 
  \And
    Lynton Ardizzone \\
  Visual Learning Lab, IWR\\
  Heidelberg University\\
  Im Neuenheimer Feld 205, 69120 Heidelberg \\
  \texttt{lynton.ardizzone@iwr.uni-heidelberg.de}
  \And
  Ullrich Köthe \\
  Visual Learning Lab, IWR\\
  Heidelberg University\\
  Im Neuenheimer Feld 205, 69120 Heidelberg\\
  \texttt{ullrich.koethe@iwr.uni-heidelberg.de}
}

\begin{document}

\date{}
\maketitle

\begin{abstract}
Estimating the parameters of mathematical models is a common problem in almost all branches of science. However, this problem can prove notably difficult when processes and model descriptions become increasingly complex and an explicit likelihood function is not available. With this work, we propose a novel method for globally amortized Bayesian inference based on invertible neural networks which we call BayesFlow. The method uses simulation to learn a global estimator for the probabilistic mapping from observed data to underlying model parameters. A neural network pre-trained in this way can then, without additional training or optimization, infer full posteriors on arbitrary many real datasets involving the same model family. In addition, our method incorporates a summary network trained to embed the observed data into maximally informative summary statistics. Learning summary statistics from data makes the method applicable to modeling scenarios where standard inference techniques with hand-crafted summary statistics fail. We demonstrate the utility of BayesFlow on challenging intractable models from population dynamics, epidemiology, cognitive science and ecology. We argue that BayesFlow provides a general framework for building amortized Bayesian parameter estimation machines for any forward model from which data can be simulated.
\end{abstract}

\section{Introduction}

The goal of Bayesian analysis is to infer the underlying characteristics of some natural process of interest given observable manifestations $\bs{x}$. In a Bayesian setting, we assume that we already posses sufficient understanding of the forward problem, that is, a suitable model of the mechanism that generates observations from a given configuration of the hidden parameters $\bs{\theta}$.
This forward model can be provided in two forms:
In likelihood-based approaches, the likelihood function $p(\bs{x}\,|\,\bs{\theta})$ is {\em explicitly known} and can be {\em evaluated} analytically or numerically for any pair $(\bs{x},\bs{\theta})$. 
In contrast, likelihood-free approaches only require the ability to {\em sample} from the likelihood. 
The latter approaches are typically realized by simulation programs, which generate synthetic observations by means of a deterministic function $g$ of parameters $\bs{\theta}$ and independent noise (i.e., random numbers) $\xib$:
\begin{equation} 
\bs{x}_i \sim p(\bs{x}\,|\,\bs{\theta}) \quad\Longleftrightarrow\quad \bs{x}_i = g(\bs{\theta},\xib_i)\,\textrm{ with }\xib_i \sim p(\xib) \label{eq:1}  
\end{equation}

In such cases, the likelihood $p(\bs{x}\,|\,\bs{\theta})$ is only defined {\em implicitly} via the action of the simulation program $g$, but calculation of its actual numerical value for a simulated observation $\bs{x}_i$ is impossible.
This, in turn, prohibits standard statistical inference.

Likelihood-free problems arise, for example, when $p(\bs{x}\,|\,\bs{\theta})$ is not available in closed-form, or when the forward model is defined by a stochastic differential equation, a Monte-Carlo simulation, or a complicated algorithm \cite{klinger2018pyabc,voss2019sequential, turner2014generalized, wood2010statistical}.
In this paper, we propose a new Bayesian solution to the likelihood-free setting in terms of \textit{invertible neural networks}.

Bayesian modeling leverages the available knowledge about the forward model to get the best possible estimate of the posterior distribution of the inverse model:
\begin{align}
    p(\bs{\theta}\,|\,\bs{x}_{1:N})=\frac{p(\bs{x}_{1:N}\,|\,\bs{\theta})\,p(\bs{\theta})}{\int p(\bs{x}_{1:N}\,|\,\bs{\theta})\,p(\bs{\theta})\,d\bs{\theta}}
\end{align}

In Bayesian inference, the posterior encodes all information about $\bs{\theta}$ obtainable from a set of observations $\bs{x}_{1:N} = \{\bs{x}_i\}_{i=1}^N$. 
The observations are assumed to arise from $N$ runs of the forward model with fixed, but unknown, true parameters $\bs{\theta}^*$. 
Bayesian inverse modeling is challenging for three reasons:

\begin{enumerate}
    \item The right-hand side of Bayes' formula above is always intractable in the likelihood-free case and must be approximated.
    \item The forward model is usually non-deterministic, so that there is intrinsic uncertainty about the true value of $\bs{\theta}$.
    \item The forward model is typically not information-preserving, so that there is ambiguity among possible values of $\bs{\theta}$.
\end{enumerate}
The standard solution to these problems is offered by \textit{approximate Bayesian computation} (ABC) methods \cite{sunnaaker2013approximate, csillery2010approximate, park2016k2,turner2014generalized}. 
ABC methods approximate the posterior by repeatedly sampling parameters from a proposal (prior) distribution $\bs{\theta}^{(l)}\sim p(\bs{\theta})$ and then simulating multiple datasets by running the forward model $\bs{x}_i\sim p(\bs{x}\,|\,\bs{\theta}^{(l)})$ for $i=1...N$. 
If the resulting dataset is sufficiently similar to the actually observed dataset $\bs{x}^{o}_{1:N}$, the corresponding $\bs{\theta}^{(l)}$ is retained as a sample from the desired posterior, otherwise rejected. 
Stricter similarity criteria lead to more accurate approximations of the desired posterior at the price of higher and oftentimes prohibitive rejection rates. 

More efficient methods for approximate inference, such as sequential Monte Carlo (ABC-SMC), Markov-Chain Monte Carlo variants \cite{sisson2011likelihood}, or the recent neural density estimation methods \cite{greenberg2019automatic, papamakarios2018sequential, lueckmann2017flexible}, optimize sampling from a proposal distribution in order to balance the speed-accuracy trade-off of vanilla ABC methods. 
More details can be found in the section \textbf{Related Work} and in the excellent review by \cite{cranmer2019frontier}.

\begin{figure*}
  \begin{subfigure}[b]{0.49\textwidth}
  	\includegraphics[width=\textwidth]{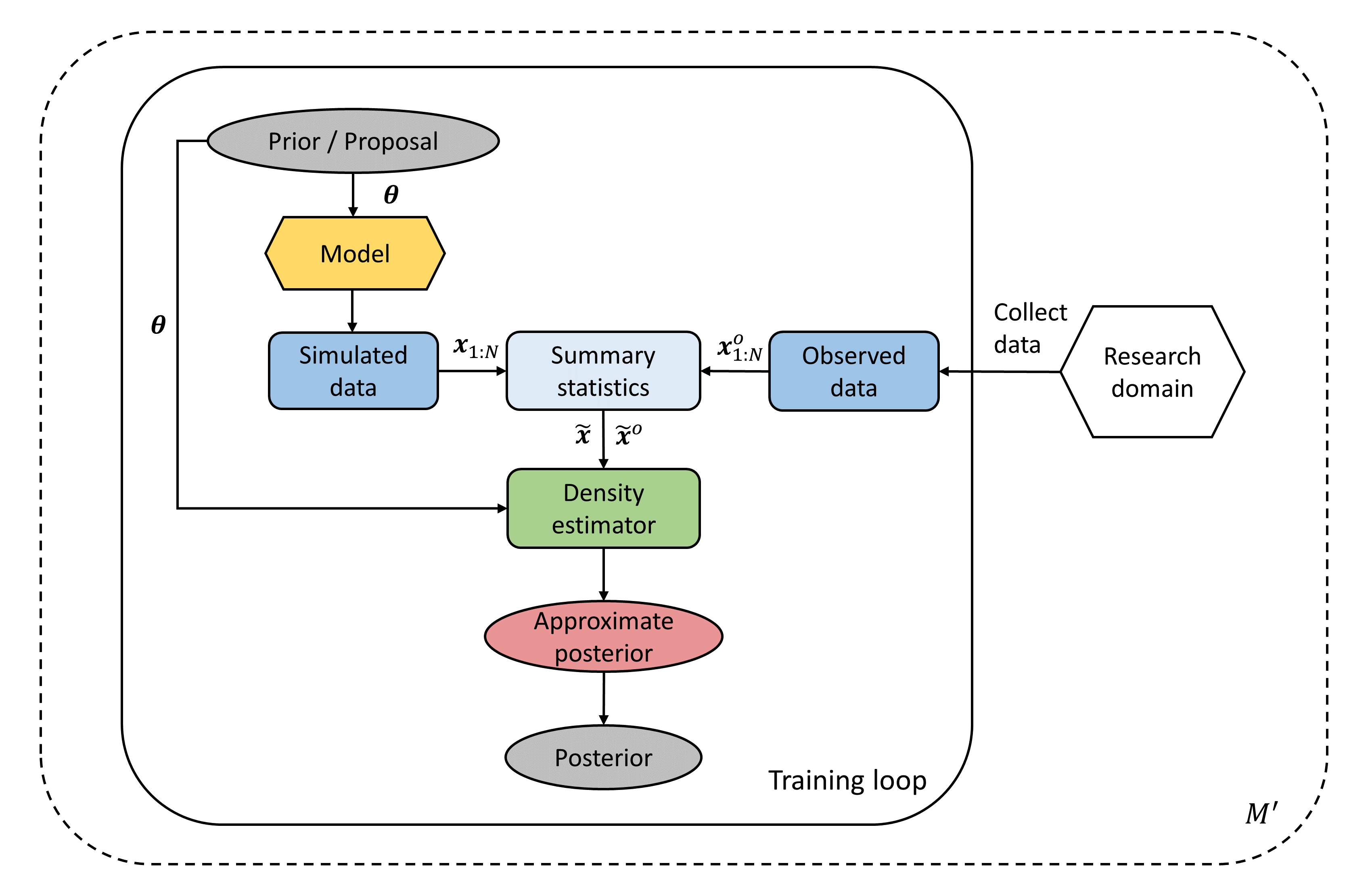}
    \caption{Case-based inference}
    \label{fig:Fig.0a}
  \end{subfigure}
  \begin{subfigure}[b]{0.49\textwidth}
  \includegraphics[width=\textwidth]{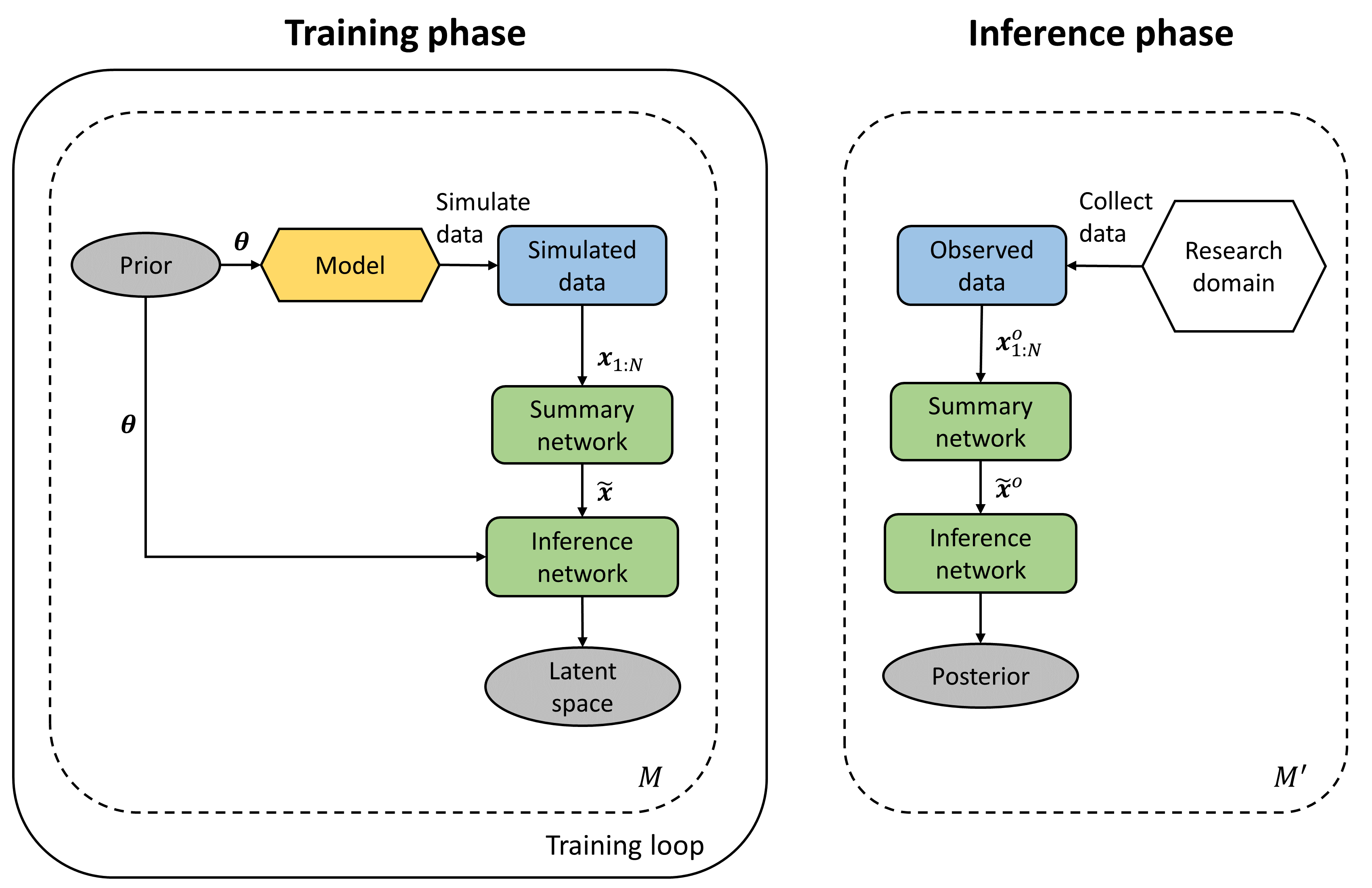}
    \caption{Globally amortized inference with BayesFlow}
    \label{fig:Fig.0b}
  \end{subfigure}
  \caption{Graphical illustration of the main differences between case-based (neural) density estimation methods and BayesFlow. (\textbf{a}) Case-based methods require a separate optimization loop for each observed dataset from a given research domain. When case-based methods incorporate a training phase (e.g., APT), it must be repeated for each new dataset. Summary statistics are manually selected and may thus be sub-optimal; (\textbf{b}) BayesFlow incorporates a global upfront training (before any real data are collected) via simulations from the forward model (left panel). Summary and inference network are trained jointly, resulting in higher accuracy than hand-crafted summary statistics. In the inference phase (right panel), BayesFlow works entirely in a feed-forward manner, that is, no training or optimization happens in this phase. The upfront training effort is therefore amortized over arbitrary many observed datasets  from a research domain working on the same model family. Note that the solid and dashed plates are swapped between case-based Bayesian inference and the training phase of BayesFlow.} \label{fig:Fig.0}
\end{figure*}

All sampling methods described above operate on the level of individual datasets, that is,
for each observation sequence $\bs{x}_{1:N}$, the entire estimation procedure for the posterior must be run again from scratch. 
Therefore, we refer to this approach as \textit{case-based inference}. 
Running estimation for each individual dataset separately stands in contrast to \textit{amortized inference}, where estimation is split into a potentially expensive \textit{upfront} training phase, followed by a much cheaper inference phase.
The goal of the upfront training phase is to learn an approximate posterior $\widehat{p}(\bs{\theta}\,|\,\bs{x}_{1:N})$ that works well for {\em any} observation sequence $\bs{x}_{1:N}$.
Evaluating this model for specific observations $\bs{x}^{o}_{1:N}$ is then very fast, so that the training effort amortizes over repeated evaluations (see \autoref{fig:Fig.0} for a graphical illustration).
The break-even between  case-based and amortized inference depends on the application and model types, and we will report comparisons in the experimental section.
Our main aim in this paper, however, is the introduction of a general approach to amortized Bayesian inference and the demonstration of its excellent accuracy in posterior estimation for a variety of popular forward models.

To make amortized inference feasible in practice, it must work well for arbitrary dataset sizes $N$. 
Depending on data acquisition circumstances, the number of available observations for a fixed model parameter setting may vary considerably, ranging from $N=1$ to several hundreds and beyond.
This has not only consequences for the required architecture of our density approximators, but also for their behavior: 
They must exhibit correct \textit{posterior contraction}. 
Accordingly, the estimated posterior $\widehat{p}(\bs{\theta}\,|\,\bs{x}_{1:N})$ should get sharper (i.e., more peaked) as the number $N$ of available observations increases. In the simplest case, the posterior variance should decrease at rate $1/N$, but more complex behavior can occur for difficult (e.g., multi-modal) true posteriors $p(\bs{\theta}\,|\,\bs{x}_{1:N})$.

We incorporate these considerations into our method by integrating two separate deep neural networks modules (detailed in the \textbf{Methods} section; see also Figure \ref{fig:Fig.0}), which are trained jointly on simulated data from the forward model: a \textit{summary network} and an \textit{inference network}.

The \textit{summary network} is responsible for reducing a set of observations $\bs{x}_{1:N}$ of variable size to a fixed-size vector of \textit{learned} summary statistics.
In traditional likelihood-free approaches, the method designer is responsible for selecting suitable statistics for each application \textit{a priori} \cite{mestdagh2019prepaid,mertens2018abrox,raynal2018abc, sunnaaker2013approximate}.
In contrast, our summary networks learn the most informative statistics directly from data, and we will show experimentally (see \textbf{Experiment 3.8}) that these statistics are superior to manually constructed ones.
Summary networks differ from standard feed-forward networks because they should be independent of the input size $N$ and respect the inherent functional and probabilistic symmetries of the data.
For example, permutation invariant networks are required for \textit{i.i.d.} observations \cite{bloem2019probabilistic}, and recurrent networks \cite{gers1999learning} or convolutional networks \cite{long2015fully} for data with temporal or spatial dependencies. 

The \textit{inference network} is responsible for learning the true posterior of model parameters given the summary statistics of the observed data. 
Since it sees the data only through the lens of the summary network, all symmetries captured by the latter are automatically inherited by the posterior.
We implement the inference network as an \textit{invertible neural network}. 
Invertible neural networks are based on the recent theory and applications of \textit{normalizing flows} \cite{ardizzone2019guided, kingma2018glow, grover2018flow, dinh2016density, kingma2016improved}. 
Flow-based methods can perform exact inference under perfect convergence and scale favourably from simple low-dimensional problems to high-dimensional distributions with complex dependencies, for instance, the pixels of an image \cite{kingma2018glow}. 
For each application/model of interest, we train an invertible network jointly with a corresponding summary network using simulated data from the respective known forward model with reasonable priors.
After convergence of this forward training, the network's invertibility ensures that a model for the inverse model is obtained for free, simply by running inference through the model backwards.
Thus, our networks can perform fast amortized Bayesian inference on arbitrary many datasets from a given application domain without expensive case-based optimization.
We call our method \textit{BayesFlow}, as it combines ideas from Bayesian inference and flow-based deep learning.

BayesFlow draws on major advances in modern deep probabilistic modeling, also referred to as deep generative modeling \cite{bloem2019probabilistic, kingma2018glow, ardizzone2018analyzing, kingma2014auto}. 
A hallmark idea in deep probabilistic modeling is to represent a complicated target distribution as a non-linear bijective transformation of some simpler latent distribution (e.g., Gaussian or uniform), a so called \textit{pushforward}.
Density estimation of the target distribution, a very complex problem, is thus reduced to learning a non-linear transformation, a task that is ideally suited for gradient-based neural network training via standard backpropagation.
During the \textit{inference phase}, samples from the target distribution are obtained by sampling from the simpler latent distribution and applying the inverse transformation learned during the training phase (see \autoref{fig:Fig.0b} for a high-level overview).
Using this approach, recent applications of deep probabilistic models have achieved unprecedented performance on hitherto intractable high-dimensional problems \cite{bloem2019probabilistic, kingma2018glow, grover2018flow}.

In the context of Bayesian inference, the target distribution is the posterior $p(\bs{\theta}\,|\,\bs{x}_{1:N})$ of model parameters given observed data. 
We leverage the fact that we can simulate arbitrarily large amounts of training data from the forward model in order to ensure that the summary and invertible networks approximate the true posterior as well as possible. 
During the inference phase, our model can either numerically evaluate the posterior probability of any candidate parameter $\bs{\theta}$,
or can generate a posterior sample $\bs{\theta}^{(1)}, \bs{\theta}^{(2)},..., \bs{\theta}^{(L)}$ of likely parameters for the observed data $\bs{x}_{1:N}^{o}$. 
In the \textbf{Methods} section, we show that our networks indeed sample from the correct posterior under perfect convergence.
In summary, the contributions of our BayesFlow method are the following:
\begin{itemize}
  \item Globally amortized approximate Bayesian inference with invertible neural networks;
  \item Learning maximally informative summary statistics from raw datasets with variable number of observations instead of relying on restrictive hand-crafted summary statistics;
  \item Theoretical guarantee for sampling from the true posterior distribution with arbitrary priors and posteriors;
  \item Parallel computations applicable to both forward simulations and neural network optimization;
\end{itemize}
To illustrate the utility of BayesFlow, we first apply it to two toy models with analytically tractable posteriors. The first is a multivariate Gaussian with a full covariance matrix and a unimodal posterior. The second is a Gaussian mixture model with a multimodal posterior. Then, we present applications to challenging models with intractable likelihoods from population dynamics, cognitive science, epidemiology, and ecology and demonstrate the utility of BayesFlow in terms of speed, accuracy of recovery, and probabilistic calibration. Alongside, we introduce several performance validation tools. 

\subsection{Related Work}

BayesFlow incorporates ideas from previous machine learning and deep learning approaches to likelihood-free inference \cite{mertens2019deep, radev2019towards, mestdagh2019prepaid, raynal2018abc, jiang2017learning}. The most common approach has been to cast the problem of parameter estimation as a supervised learning task. 
In this setting, a large dataset of the form $\bs{D} = \{(h(\bs{x}_{1:N}^{(m)}), \bs{\theta}^{(m)})\}_{m=1}^{M}$ is created by repeatedly sampling from $p(\bs{\theta})$ and simulating an artificial datasets $\bs{x}_{1:N}$ by running the simulator with the sampled parameters. 
Usually, the dimensionality of the simulated data is reduced by computing summary statistics  with a fixed summary function $h(\bs{x}_{1:N})$. 
Then, a supervised learning algorithm (e.g., random forest  \cite{raynal2018abc}, or a neural network \cite{radev2019towards}) is trained on the summary statistics of the simulated data to output an estimate of the true data generating parameters. 
Thus, an attempt is made to approximate the intractable inverse model $\bs{\theta} = g^{-1}(\bs{x},\bs{\xi})$. 
A main shortcoming of supervised approaches is that they provide only limited information about the posterior (e.g., point-estimates, quantiles or variance estimates) or impose overly restrictive distributional assumptions on the shape of the posterior (e.g., Gaussian).

Our ideas are also closely related to the concept of \textit{optimal transport maps} and its application in Bayesian inference \cite{detommaso2019hint,parno2018transport,chen2018fast,bigoni2019greedy}. 
A transport map defines a transformation between (probability) measures which can be constructed in a way to \textit{warp} a simple probability distribution into a more complex one. 
In the context of Bayesian inference, transport maps have been applied to accelerate MCMC sampling \cite{parno2018transport}, to perform sequential inference \cite{detommaso2019hint}, and to solve inference problems via direct optimization \cite{bigoni2019greedy}. 
In fact, BayesFlow can be viewed as a parameterization of invertible transport maps via invertible neural networks. 
An important distinction is that BayesFlow does not require an explicit likelihood function for approximating the target posteriors and is capable of amortized inference. 

Similar ideas for likelihood-free inference are incorporated in the recent automatic posterior transformation (APT) \cite{greenberg2019automatic}, and the sequential neural likelihood (SNL) \cite{papamakarios2018sequential} methods. APT iteratively refines a proposal distribution via masked autoregressive flow (MAF) networks to generate parameter samples which closely match a particular observed dataset. SNL, in turn, trains a masked autoencoder density estimator (MADE) neural network within an MCMC loop to speed-up convergence to the true posterior. 
Even though these methods also entail a relatively expensive learning phase and a cheap inference phase, posterior inference is amortized only for a single dataset.
Thus, the learning phase needs to be run through for each individual dataset (see \autoref{fig:Fig.0a}).
In contrast, we propose to learn the posterior globally over the entire range of plausible parameters and datasets by employing a conditional invertible neural network (cINN) estimator (see \autoref{fig:Fig.0b}). 
Previously, INNs have been successfully employed to model data from astrophysics and medicine \cite{ardizzone2018analyzing}. 
We adapt the model to suit the task of parameter estimation in the context of mathematical modeling and develop a probabilistic architecture for performing fully Bayesian and globally amortized inference with complex mathematical models.

\section{Methods}

\subsection{Notation}

In the following, the number of parameters of a mathematical model will be denoted as $D$, and the number of observations in a dataset as $N$. We denote data simulated from the mathematical model of interest as $\bs{x}_{1:N} = (\bs{x}_{1}, \bs{x}_{2},...,\bs{x}_{N})$, where each individual $\bs{x}_{i}$ can represent a scalar or a vector. Observed or test data will be marked with a superscript $o$ (i.e., $\bs{x}_{1:N}^{o}$). The parameters of a mathematical model are represented as a vector $\bs{\theta} = (\theta_{1}, \theta_{2},...,\theta_{D})$, and all trainable parameters of the invertible and summary neural networks as $\bs{\phi}$ and $\bs{\psi}$, respectively. When a dataset consists of observations over a period of time, the number of observations will be denotes as $T$. 

\subsection{Learning the Posterior}

Assume that we have an invertible function $f_{\bs{\phi}}: \mathbb{R}^{D} \rightarrow \mathbb{R}^{D}$, parameterized by a vector of parameters $\bs{\phi}$, for which the inverse $f_{\bs{\phi}}^{-1}: \mathbb{R}^{D} \rightarrow \mathbb{R}^{D}$ exists. For now, consider the case when raw simulated data $\bs{x}_{1:N}$ of size $N=1$ is entered directly into the invertible network without using a summary network. Our goal is to train an invertible neural network which approximates the true posterior as accurately as possible:
\begin{align} 
p_{\bs{\phi}}(\bs{\theta}\,|\,\bs{x}) \approx p(\bs{\theta}\,|\,\bs{x}) \label{eq:7}
\end{align}
for all possible $\bs{\theta}$ and $\bs{x}$. We reparameterize the approximate posterior $p_{\bs{\phi}}$ in terms of a conditional invertible neural network (cINN) $f_{\bs{\phi}}$ which implements a normalizing flow between $\bs{\theta}$ and a Gaussian latent variable $\bs{z}$:
\begin{align}
\bs{\theta} \sim p_{\bs{\phi}}(\bs{\theta}\,|\,\bs{x}) \Longleftrightarrow \bs{\theta} = f^{-1}_{\bs{\phi}}(\bs{z};\bs{x}) \textrm{ with } \bs{z} \sim \mathcal{N}_{D}(\bs{z}\,|\,\bs{0},\bs{\mathbb{I}}) \label{eq:8}
\end{align}
Accordingly, we need to ensure that the outputs of $f_{\bs{\phi}}^{-1}(\bs{z};\bs{x})$ follow the target posterior $p(\bs{\theta}\,|\,\bs{x})$. Thus, we seek neural network parameters $\widehat{\bs{\phi}}$ which minimize the Kullback-Leibler (KL) divergence between the true and the model-induced posterior for all possible datasets $\bs{x}$. Therefore, our objective becomes:
\begin{align}
\widehat{\bs{\phi}}
&= \argmin_{\bs{\phi}}\mathbb{E}_{ p(\bs{x})}\left[\mathbb{KL}(p(\bs{\theta}\,|\,\bs{x})\,||\,p_{\bs{\phi}}(\bs{\theta}\,|\,\bs{x}))\right] \label{eq:9} \\
&= \argmin_{\bs{\phi}}\mathbb{E}_{ p(\bs{x})}\left[\mathbb{E}_{ p(\bs{\theta}\,|\,\bs{x})}\left[\log p(\bs{\theta}\,|\,\bs{x}) - \log p_{\bs{\phi}}(\bs{\theta}\,|\,\bs{x})\right]\right]  \label{eq:10}  \\
&= \argmax_{\bs{\phi}}\mathbb{E}_{ p(\bs{x})}\left[\mathbb{E}_{ p(\bs{\theta}\,|\,\bs{x})}\left[\log p_{\bs{\phi}}(\bs{\theta}\,|\,\bs{x})\right]\right]  \label{eq:11} \\
&= \argmax_{\bs{\phi}} \int\!\!\!\int p(\bs{x},\bs{\theta}) \log p_{\bs{\phi}}(\bs{\theta}\,|\,\bs{x})d\bs{x}d\bs{\theta} \label{eq:12}
\end{align}
Note, that the log posterior density $p(\bs{\theta}\,|\,\bs{x})$ can be dropped from the optimization objective in Eq.\ref{eq:11}, as it does not depend on the neural network parameters $\bs{\phi}$. In other words, we seek neural network parameters $\widehat{\bs{\phi}}$ which maximize the posterior probability of data-generating parameters $\bs{\theta}$ given observed data $\bs{x}$ for all $\bs{\theta}$ and $\bs{x}$. Since $f_{\bs{\phi}}(\bs{\theta};\bs{x}) = \bs{z}$ by design, the change of variable rule of probability yields:
\begin{align}
p_{\bs{\phi}}(\bs{\theta}\,|\,\bs{x}) = p(\bs{z}=f_{\bs{\phi}}(\bs{\theta};\bs{x}))\left|\det \left(\frac{\partial f_{\bs{\phi}}(\bs{\theta};\bs{x})}{\partial \bs{\theta}}\right)\right| \label{eq:13}
\end{align} 
Thus, we can re-write our objective as:
\begin{align}
\widehat{\bs{\phi}} 
&= \argmax_{\bs{\phi}}\int\!\!\!\int p(\bs{x},\bs{\theta})\log p_{\bs{\phi}}(\bs{\theta}\,|\,\bs{x}) d\bs{x} d\bs{\theta} \label{eq:14} \\ 
&= \argmax_{\bs{\phi}}\int\!\!\!\int  p(\bs{x},\bs{\theta}) \big( \log p(f_{\bs{\phi}}(\bs{\theta};\bs{x})) + \log \left| \det \bs{J}_{f_{\bs{\phi}}} \right| \big) d\bs{x} d\bs{\theta} \label{eq:15}
\end{align}
where we have abbreviated $\partial f_{\bs{\phi}}(\bs{\theta};\bs{x}) / \partial \bs{\theta}$ (the Jacobian of $f_{\bs{\phi}}$ evaluated at $\bs{\theta}$ and $\bs{x}$) as $\bs{J}_{f_{\bs{\phi}}}$. Due to the architecture of our cINN, the $\log \left| \det \bs{J}_{f_{\bs{\phi}}} \right|$ is easy to compute (see next section for details).

Utilizing simulations from the forward model (Eq.\ref{eq:1}), we can approximate the expectations by minimizing the Monte-Carlo estimate of the negative of Eq.\ref{eq:15}. Accordingly, for a batch of $M$ simulated datasets and data-generating parameters $\{(\bs{x}^{(m)}, \bs{\theta}^{(m)})\}_{m=1}^M$ we have:
\begin{align}
\widehat{\bs{\phi}} 
&= \argmin_{\bs{\phi}} 
 \frac{1}M\sum_{m=1}^M
-\log p_{\bs{\phi}}(\bs{\theta}^{(m)}\,|\,\bs{x}^{(m)})  \label{eq:16} \\
&= \argmin_{\bs{\phi}} 
 \frac{1}M\sum_{m=1}^M\left(-\log p(f_{\bs{\phi}}(\bs{\theta}^{(m)};\bs{x}^{(m)})) - \log \left| \det \bs{J}^{(m)}_{f_{\bs{\phi}}} \right| \right)   \label{eq:17} \\
&= \argmin_{\bs{\phi}} \frac{1}M\sum_{m=1}^M\left(\frac{\norm{f_{\bs{\phi}}(\bs{\theta}^{(m)};\bs{x}^{(m)})}_{2}^{2}}{2} - \log\left|\det \bs{J}_{f_{\bs{\phi}}}^{(m)}\right|\right) \label{eq:18}
\end{align}

We treat Eq.\ref{eq:18} as a loss function $\mathcal{L}(\bs{\phi})$ which can be minimized with any stochastic gradient descent method. The first term follows from Eq.\ref{eq:17} due to the fact that we have prescribed a unit Gaussian distribution to $\bs{z}$. It represents the negative log of $\mathcal{N}_{D}(\bs{z}\,|\,\bs{0},\bs{\mathbb{I}}) \propto \exp(\norm{-\frac{1}{2}\bs{z}}_{2}^{2})$. The second term controls the rate of volume change induced by the learned non-linear transformation from $\bs{\theta}$ to $\bs{z}$ achieved by $f_{\bs{\phi}}$. Thus, minimizing Eq.\ref{eq:18} ensures that $\bs{z}$ follows the prescribed unit Gaussian. 

The correctness of the learned posterior can be guaranteed in the following way,
assuming the network is able to reach the global minimum of the loss (i.e. under perfect convergence).
\newcommand{\z}{\bs{z}}
\newcommand{\x}{\bs{x}}
\newcommand{\thh}{\bs{\theta}}
\newcommand{\INN}{f_{\bs{\widehat{\phi}}}}
\newcommand{\pphi}{p_{\bs{\hat \phi}}}

\begin{prop}
Assume that the cINN architecture and domain of $\phi$ are chosen such that $\widehat{\phi}$ is the global minimum of the objective in Eq.\ref{eq:15}.
Then, the latent output distribution will be statistically independent of the conditioning data, $\pphi(\z \mid \x) \perp p(\x)$.
As a result, the samples transformed backwards from $p(\z)$ will follow the true posterior, that is:
\begin{equation}
    \INN^{-1}(\bs{z}; \bs{x}) \sim p(\bs{\theta} \mid \bs{x}) \quad \text{with} \quad \bs{z} \sim \mathcal{N}_{D}(\bs{z}\,|\,\bs{0},\bs{\mathbb{I}})
\end{equation}
\end{prop}
\begin{proof}

For short, we denote $p(\z) \coloneqq \mathcal{N}_{D}(\bs{z}\,|\,\bs{0},\bs{\mathbb{I}})$,
and the distribution of network outputs as $p(\INN(\thh; \x)) \coloneqq \pphi ( \z \mid \x)$.
Due to $\mathbb{KL}(\cdot \| \cdot) \geq 0$ (Gibbs' inequality), the global minimum of the objective is achieved
exactly when the argument in Eq.\ref{eq:9} becomes 0.
To relate this to the sampling process,
we note the invariance of $\mathbb{KL}$ under diffeomorphic transformations, from which it follows that
\begin{equation}
\mathbb{KL}\left( \pphi(\z \mid \x)  \,\|\, p(\z) \right) = 0.\label{eq:latent_kl}
\end{equation}
Considering $p(\z) \perp p(\x)$ and $\pphi(\z\,|\,\x) = p(\z)$ (from Eq.\ref{eq:latent_kl}), this also implies $\pphi(\z\,|\,\x) \perp p(\x)$,
which means the latent output distribution is the same for any fixed $\x$ we choose.
This motivates the validity of taking samples from $p(\z)$ and transforming them back using the condition, 
to generate samples from the posterior.
%
%
By definition of the model, the generated samples $\INN^{-1}(\z, \x)$ with $\z \sim p(\z)$ 
follow $\pphi(\thh\,|\,\x)$.
The proposition therefore holds when the argument in Eq.\ref{eq:9} is zero.
\end{proof}

We now generalize our formulation to datasets with arbitrary numbers of observations. If we let the number of observations $N$ vary and train a summary network $\tilde{\bs{x}} = h_{\bs{\psi}}(\bs{x}_{1:N})$ together with the cINN, our main objective changes to:
\begin{align}
\widehat{\bs{\phi}},\widehat{\bs{\psi}} = \argmax_{\bs{\phi},\bs{\psi}}\mathbb{E}_{ p(\bs{x},\bs{\theta},N)}\left[\log p_{\bs{\phi}}(\bs{\theta}\,|\,h_{\bs{\psi}}(\bs{x}_{1:N}))\right] \label{eq:24}
\end{align}
and its Monte Carlo estimate to:
\begin{equation}
\begin{split}
\widehat{\bs{\phi}},\widehat{\bs{\psi}} = \argmin_{\bs{\phi},\bs{\psi}} \frac{1}M\sum_{m=1}^M \left( \frac{\norm{f_{\bs{\phi}}(\bs{\theta}^{(m)};h_{\bs{\psi}}(\bs{x}^{(m)}_{1:N})}_{2}^{2}}{2}  -  \log\left|\det\left(\bs{J}_{f_{\bs{\phi}}}^{(m)}\right)\right| \right) \label{eq:25}
\end{split}
\end{equation}
In order to make the estimation of $p(\bs{\theta}\,|\,\bs{x}_{1:N})$ tractable, we assume that there exists a vector $\bs{\eta}$ of sufficient statistics that captures all information about $\bs{\theta}$ contained in $\bs{x}_{1:N}$ in a fixed-size (vector) representation. For $h_{\bs{\psi}}(\bs{x}_{1:N})$ to be a useful estimator for $\bs{\eta}$, both should convey the same information about $\bs{\theta}$, as measured by the mutual information:
\begin{align}
MI(\bs{\theta},h_{\bs{\psi}}(\bs{x}_{1:N})) \approx MI(\bs{\theta},\bs{\eta}) \label{eq:26}
\end{align}

Since we do not know $\bs{\eta}$, we can enforce this requirement only indirectly by minimizing the Monte Carlo estimate of Eq.\ref{eq:24}. The following proposition states that, under perfect convergence, samples from a cINN still follow the true posterior given the outputs of a summary networks. 

\begin{prop}
Assume that we have a perfectly converged cINN $f_{\bs{\phi}}$ and a perfectly converged summary network $h_{\bs{\psi}}$. Assume also, that there exists a vector $\bs{\eta}$ of sufficient summary statistics for $\bs{x}_{1:N}$. Then, independently sampling $\bs{z} \sim p(\bs{z})$ and applying $f_{\bs{\phi}}^{-1}(\bs{z};h_{\bs{\psi}}(\bs{x}_{1:N}))$ to each $\bs{z}$ yields independent samples from $p(\bs{\theta}\,|\,\bs{x}_{1:N})$.
\end{prop}

\begin{proof}
Perfect convergence of the networks under Eq.\ref{eq:24} implies $\mathbb{KL}(p(\bs{\theta}\,|\,\bs{x}_{1:N})\,||\,p_{\bs{\phi}}(\bs{\theta}\,|\,h_{\bs{\psi}}(\bs{x}_{1:N}))) = 0$. This, in turn, implies that $MI(\bs{\theta},h_{\bs{\psi}}(\bs{x}_{1:N})) = MI(\bs{\theta},\bs{\eta})$, because a perfect match of the densities would be impossible if $h_{\bs{\psi}}(\bs{x}_{1:N})$ contained less information about $\bs{\theta}$ than $\bs{\eta}$. Therefore, the proof reduces to that of \textbf{Proposition 1}. Note, that whenever the KL divergence is driven to a minimum, $h_{\bs{\psi}}(\bs{x}_{1:N})$ is a \textit{maximally informative statistic} \cite{deans2002maximally}.
\end{proof}

In summary, the approximate posteriors obtained by the BayesFlow method are correct if the summary and invertible networks are perfectly converged. In practice, however, perfect convergence is unrealistic and there are three sources of error which can lead to incorrect posteriors. The first is the Monte Carlo error introduced by using simulations from $g(\bs{\theta},\xib)$ to approximate the expectation in Eq.\ref{eq:24}. The second is due to a summary network which may not fully capture the relevant information in the data or when sufficient summary statistics do not exist. The third is due to an invertible network which does not accurately transform the true posterior into the prescribed Gaussian latent space.
Even though we can mitigate the Monte Carlo error by running the simulator $g(\bs{\theta},\bs{\xi})$ more often, the latter two can be harder to detect and alleviate in a principled way. Nevertheless, recent work on \textit{probabilistic symmetry} \cite{bloem2019probabilistic} and \textit{algorithmic alignment} \cite{xu2019what} can provide some guidelines on how to choose the right summary network for a particular problem. Additionally, the depth as well as the building blocks (to be explained shortly) of the invertible chain can be tuned to increase the expressiveness of the learned transformation from $\bs{\theta}$-space to $\bs{z}$-space. The benefits of neural network depth have been confirmed both in theory and practice \cite{lin2018resnet,belkin2019reconciling}, so we expect better performance in complex settings with increasing network depth.

\subsection{Composing Invertible Networks}

The basic building block of our cINN is the affine coupling block (ACB) \cite{dinh2016density}. Each ACB consists of four separate fully connected neural networks denoted as $s_{1}(\cdot), s_{2}(\cdot), t_{1}(\cdot), t_{2}(\cdot)$. An ACB performs an invertible non-linear transformation, which means that in addition to a parametric mapping $f_{\bs{\phi}}: \mathbb{R}^{D} \rightarrow \mathbb{R}^{D}$ it also learns the inverse mapping $f_{\bs{\phi}}^{-1}: \mathbb{R}^{D} \rightarrow \mathbb{R}^{D}$ \textit{for free}. Denoting the input vector of $f_{\bs{\phi}}$ as $\bs{u}$ and the output vector as $\bs{v}$, it follows that $f_{\bs{\phi}}(\bs{u}) = \bs{v}$ and $f_{\bs{\phi}}^{-1}(\bs{v}) = \bs{u}$. Invertibility is achieved by splitting the input vector into two parts $\bs{u} = (\bs{u}_{1}, \bs{u}_{2})$ with $\bs{u}_1 = u_{1:D/2}$ and $\bs{u}_2 = u_{D/2+1:D}$ (where $D/2$ is understood as a floor division) and performing the following operations on the split input:
\begin{align} 
\bs{v}_{1} &= \bs{u}_{1} \odot \exp(s_{1}(\bs{u}_{2})) + t_{1}(\bs{u}_{2}) \label{eq:3} \\ 
\bs{v}_{2} &= \bs{u}_{2} \odot \exp(s_{2}(\bs{v}_{1})) + t_{2}(\bs{v}_{1}) \label{eq:4} 
\end{align}

where $\odot$ denotes element-wise multiplication. The outputs $\bs{v} = (\bs{v}_{1}, \bs{v}_{2})$ are then concatenated again and passed to the next ACB. The inverse operation is given by:
\begin{align} 
\bs{u}_{2} &= (\bs{v}_{2} - t_{2}(\bs{v}_{1})) \odot \exp(-s_{2}(\bs{v}_{1})) \label{eq:5}  \\ 
\bs{u}_{1} &= (\bs{v}_{1} - t_{1}(\bs{u}_{2})) \odot \exp(-s_{1}(\bs{u}_{2})) \label{eq:6} 
\end{align}

This formulation ensures that the Jacobian of the affine transformation is a strictly upper or a lower triangular matrix and therefore its determinant is very cheap to compute. Furthermore, the internal functions $s_{1}(\cdot), s_{2}(\cdot), t_{1}(\cdot), t_{2}(\cdot)$ can be represented by arbitrarily complex neural networks, which themselves need not be invertible, since they are only ever evaluated in the forward direction during both the forward and the inverse pass through the ACBs. In our applications, we parameterize the internal functions as fully connected neural networks with exponential linear units (ELU).

In order to ensure that the neural network architecture is expressive enough to represent complex distributions, we chain multiple ACBs, so that the output of each ACB becomes the input to the next one. In this way, the whole chain remains invertible from the first input to the last output and can be viewed as a single function parameterized by trainable parameters $\bs{\phi}$.

In our applications, the input to the first ACB is the parameter vector $\bs{\theta}$, and the output of the final ACB is a $d$-dimensional vector $\bs{z}$ representing the non-linear transformation of the parameters. As described in the previous section, we ensure that $\bs{z}$ follows a unit Gaussian distribution via optimization, that is, $p(\bs{z}) = \mathcal{N}_{D}(\bs{z}\,|\,\bs{0},\bs{\mathbb{I}})$. Fixed permutation matrices are used before each ACB to ensure that each axis of the transformed parameter space $\bs{z}$ encodes information from all components of $\bs{\theta}$. 

In order to account for the observed data, we feed the learned summary vectors into all internal networks of each ACB (explained shortly). Intuitively, in this way we realize the following process: the forward pass maps data-generating parameters $\bs{\theta}$ to $\bs{z}$-space using conditional information from the data $\bs{x}_{1:N}$, while the inverse pass maps data points from $\bs{z}$-space to the data-generating parameters of interest using the same conditional information.

\subsection{Summary Network}

Since the number of observations usually varies in practical scenarios (e.g., different number of measurements or time points) and since datasets might exhibit various redundancies, the cINN can profit from some form of dimensionality reduction. 
As previously mentioned, we want to avoid information loss through restrictive hand-crafted summary statistics and, instead, learn the most informative summary statistics directly from data. 
Therefore, instead of feeding the raw simulated or observed data to each ACB, we pass the data through an additional summary network to obtain a fixed-sized vector of learned summary statistics $\tilde{\bs{x}} = h_{\bs{\psi}}(\bs{x}_{1:N})$. 

The architecture of the summary network should be aligned with the probabilistic symmetry of the observed data. An obvious choice for time series-data is an LSTM-network \cite{gers1999learning}, since recurrent networks can naturally deal with long sequences of variable size. Another choice might be a 1D fully convolutional network \cite{long2015fully}, which has already been applied in the context of likelihood-free inference \cite{radev2019towards}. A different architecture is needed when dealing with \textit{i.i.d.} samples of variable size. Such data are often referred to as \textit{exchangeable}, or \textit{permutation invariant}, since changing the order of individual elements does not change the associated likelihood or posterior. In other words, if $\mathbb{S}_{N}(\cdot)$ is an arbitrary permutation of $N$ elements, the following should hold for the posterior:
\begin{align}
    p(\bs{\theta}\,|\,\bs{x}_{1:N}) = p(\bs{\theta}\,|\,\mathbb{S}_{N}(\bs{x}_{1:N}))
\end{align}

Following \cite{bloem2019probabilistic}, we encode probabilistic permutation invariance by implementing a permutation invariant function through an equivariant non-linear transformation followed by a pooling operator (e.g., sum or mean) and another non-linear transformation:
\begin{align}
\tilde{\bs{x}} = h_{\bs{\psi}_1}\left(\sum_{i=1}^N h_{\bs{\psi}_2}(\bs{x}_i)\right)
\end{align}

where $h_{\bs{\psi}_1}$ and $h_{\bs{\psi}_2}$ are two different fully connected neural networks. In practice, we stack multiple equivariant and invariant functions into an invariant network in order to achieve higher expressiveness \cite{bloem2019probabilistic}. 

We optimize the parameters $\bs{\psi}$ of the summary network jointly with those of the cINN chain via backpropagation. Thus, training remains completely end-to-end, and BayesFlow learns to generalize to datasets of different sizes by suitably varying $N$ during training of a permutation invariant summary network or varying sequence length during training of a recurrent/convolutional network.

To incorporate the observed or simulated data $\bs{x}_{1:N}$, each of the internal networks of each ACB is augmented to take the learned summary vector $\tilde{\bs{x}}$ of the data as an additional input. The output of each ACB then becomes:
\begin{align} 
\bs{v}_{1} &= \bs{u}_{1} \odot \exp(s_{1}(\bs{u}_{2},\tilde{\bs{x}})) + t_{1}(\bs{u}_{2},\tilde{\bs{x}}) \\ 
\bs{v}_{2} &= \bs{u}_{2} \odot \exp(s_{2}(\bs{v}_{1},\tilde{\bs{x}})) + t_{2}(\bs{v}_{1},\tilde{\bs{x}}) 
\end{align}

Thus, a complete pass through the entire conditional invertible chain can be expressed as $f_{\bs{\phi}}(\bs{\theta};\tilde{\bs{x}}) = \bs{z}$ together with the inverse operation $f_{\bs{\phi}}^{-1}(\bs{z};\tilde{\bs{x}}) = \bs{\theta}$. The inverse transformation during inference is depicted in \autoref{fig:Fig.1}.

\subsection{Putting It All Together}

\textbf{Algorithm} \ref{alg:1} describes the essential steps of the BayesFlow method using an arbitrary summary network and employing an online learning approach.

\begin{algorithm*}
\caption{Amortized Bayesian inference with the BayesFlow method}\label{alg:1}
\begin{algorithmic}[1]
\State \emph{{\bf Training phase} (online learning with batch size $M$):}
\Repeat
\State {Sample number of observations $N \sim \mathcal{U}(N_{min}, N_{max})$.}
\For{$m = 1,...,M$} 
    \State {Sample model parameters from prior: $\bs{\theta}^{(m)} \sim p(\bs{\theta})$.}
    \For{$i = 1,...,N$} 
        \State {Sample a noise instance: $\xib_i\sim p(\xib)$.}
        \State {Run the simulation (cf. Eq.\ref{eq:1}) to create a synthetic observation: $\bs{x}_{i}^{(m)} = g(\bs{\theta}^{(m)},\xib_i)$.}
    \EndFor
    \State {Pass the dataset $\bs{x}_{1:N}^{(m)}$ through the summary network: $\widetilde{\bs{x}}^{(m)}= h_{\bs{\psi}}(\bs{x}_{1:N}^{(m)})$.}
    \State {Pass $(\bs{\theta}^{(m)},\widetilde{\bs{x}}^{(m)})$  through the inference network in forward direction: $\bs{z}^{(m)}=f_{\bs{\phi}}(\bs{\theta}^{(m)};\widetilde{\bs{x}}^{(m)})$.}
\EndFor
\State {Compute loss according to Eq.\ref{eq:25} from the training batch $\big\{(\bs{\theta}^{(m)},\widetilde{\bs{x}}^{(m)},\bs{z}^{(m)})\big\}_{m=1}^M$.}
\State {Update neural network parameters $\bs{\phi},\bs{\psi}$ via backpropagation.}
\Until{convergence to $\widehat{\bs{\phi}},\widehat{\bs{\psi}}$}\\
\State \emph{{\bf Inference phase} (given observed or test data $\bs{x}^{o}_{1:N}$):}
\State {Pass the observed dataset through the summary network: $\widetilde{\bs{x}}^{o} = h_{\widehat{\bs{\psi}}}(\bs{x}_{1:N}^{o})$}.
\For{$l = 1,...,L$}
\State {Sample a latent variable instance: $\bs{z}^{(l)} \sim \mathcal{N}_{D}(\bs{0},\bs{\mathbb{I}})$.}
\State {Pass $(\widetilde{\bs{x}}^{o},\bs{z}^{(l)})$ through the inference network in inverse direction: $\bs{\theta}^{(l)} = f_{\widehat{\bs{\phi}}}^{-1}(\bs{z}^{(l)};\widetilde{\bs{x}}^{o})$.}
\EndFor
\State {Return $\big\{\bs{\theta}^{(l)}\big\}_{l=1}^{L}$ as a sample from $p(\bs{\theta}\,|\,\bs{x}^{o}_{1:N})$}
\end{algorithmic}
\end{algorithm*}

\begin{figure*}
\centering
\begin{subfigure}{.95\textwidth}
	\includegraphics[width=\textwidth]{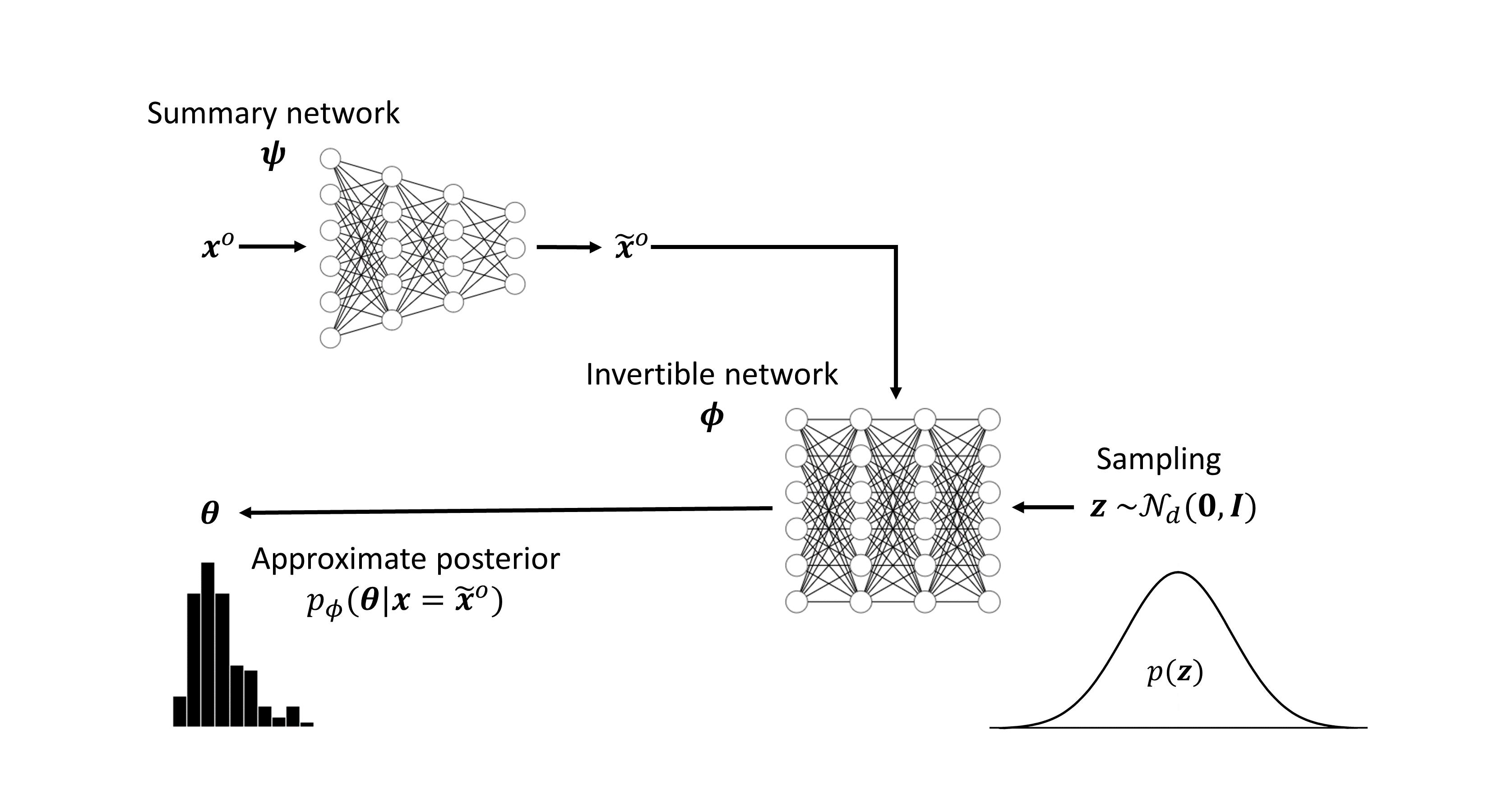}
\end{subfigure}
\caption{Inference with pre-trained summary and inference networks. The posterior is approximated given real observed data via independent samples from a learned pushforward distribution. Thus, knowledge about the mapping between data and parameters (the inverse model) is compactly encoded within the weights of the two networks.}
\label{fig:Fig.1}
\end{figure*}

The backpropagation algorithm works by computing the gradients of the loss function with respect to the parameters of the neural networks and then adjusting the parameters, so as to drive the loss function to a minimum. We experienced no instability or convergence issues during training with the loss function given by Eq.\ref{eq:25}. Note, that steps 3-14 and 18-22 of \textbf{Algorithm} \ref{alg:1} can be executed in parallel with GPU support in order to dramatically accelerate convergence and inference. Moreover, steps 18-22 can be applied in parallel to an arbitrary number of observed datasets after convergence of the networks (see \autoref{fig:Fig.1} for a full graphical illustration).

In what follows, we apply BayesFlow to two toy models with a unimodal and multimodal posteriors, respectively, and then use it to perform Bayesian inference on challenging models from population dynamics, cognitive science, epidemiology, and ecology.\footnote{Code and simulation scripts for all current applications are available at \url{https://github.com/stefanradev93/cINN}.} We deem these models suitable for an initial validation, since they differ widely in the generative mechanisms they implement and the observed data they aim to explain. Therefore, good performance on these disparate examples underlines the broad empirical utility of the BayesFlow method. Details for models' setup can be found in \textbf{Appendix C}.

\section{Experiments}

\subsection{Training the Networks}

We train all invertible and summary networks described in this paper jointly via backpropagation. For all following experiments, we use the Adam optimizer with a starter learning rate of $10^{-3}$ and an exponential decay rate of $.95$. We perform $50\ 000$ to $100\ 000$ iterations (i.e., mini-batch update steps) for each experiment, and report the results obtained by the converged networks. 
Note, that we did not perform an extensive search for optimal values of network hyperparameters, but use a default BayesFlow with 5 to 10 ACBs and a summary vector of size $128$ for all examples in this paper (see \textbf{Appendix C} for more details on summary network architectures). 
All networks were implemented in Python using the \textit{TensorFlow} library \cite{abadi2016tensorflow} and trained on a single-GPU machine equipped with NVIDIA\textsuperscript{\textregistered} GTX1060 graphics card. 

Regarding the data generation step, we take an approach which incorporates ideas from \textit{online learning} \cite{mnih2015human} where data are simulated by Eq.\ref{eq:1} on demand. 
Correspondingly, a dataset $\bs{x}_{1:N}$, or a batch of $M$ datasets $\{\bs{x}_{1:N}^{(m)}\}_{m=1}^M$, is generated on the fly and then passed through the neural network. This training approach has the advantage that the network never \textit{experiences} the same input data twice. Moreover, training can continue as long as the network keeps improving (i.e., the loss keeps decreasing), since overfitting in the classical sense is nearly impossible. However, if simulations are computationally expensive and researchers need to experiment with different networks or training hyperparameters, it might be beneficial to store and re-use simulations, since simulation and training in online learning are tightly intertwined.

Once the networks have converged, we store the trained networks and use them to perform amortized inference on a separate validation set of datasets. The pre-trained networks can also be shared among a research community so that multiple researchers/labs can benefit from the amortization of inference.

\subsection{Performance Validation}

To evaluate the performance of BayesFlow in the following application examples, we consider a number of different metrics:
\begin{itemize}
    \item Normalized root mean squared error (NRMSE) - to asses accuracy of point-estimates in recovering ground-truth parameter values;
    \item Coefficient of determination ($R^{2}$) - to assess the proportion of variance in ground-truth parameters that is captured by the point estimates; 
    \item Re-simulation error ($Err_{sim}$) - to assess the predictive mismatch between the true data distribution and the data distribution generated with the estimated parameters (i.e., posterior predictive check);
    \item Calibration error ($Err_{cal}$, \cite{ardizzone2018analyzing}) - to assess the coverage of the approximate posteriors (i.e., whether credibility intervals are indeed credible); 
    \item Simulation-based calibration (SBC, \cite{talts2018validating}) - to visually detect systematic biases in the approximate posteriors;
\end{itemize}
Details for computing all metrics are given in \textbf{Appendix B}. 

\subsection{Proof of Concept: Multivariate Normal Distribution}

As a proof-of-concept, we apply the BayesFlow method to recover the posterior mean vector of a toy multivariate normal (MVN) example. For a single $D$-dimensional MVN vector, the forward model is given by:
\begin{align}
\bs{\mu}^{(m)} &\sim \mathcal{N}_{D}(\bs{\mu}\,|\,\bs{0},\bs{\mathbb{I}})  \\
\bs{x}^{(m)} &\sim \mathcal{N}_{D}(\bs{x}\,|\,\bs{\mu}^{(m)},\bs{\Sigma}) 
\end{align}
where in this illustrative case we assume a single $D$-dimensional sample per observation ($N=1$). If the covariance matrix $\bs{\Sigma}$ is known, the posterior of the mean vector $\bs{\mu}$ has a closed-form which is also a MVN $p(\bs{\mu}\,|\,\bs{x},\bs{\Sigma}) = \mathcal{N}_{d}(\bs{\mu}\,|\,\bs{m},\bs{\Lambda})$ with posterior precision matrix given by $\bs{\Lambda}^{-1} = \bs{\mathbb{I}} + \bs{\Sigma}^{-1}$ and posterior mean given by $\bs{m} = \bs{\Lambda}\bs{\Sigma}^{-1}\bs{x}$ \cite{bolstad2016introduction}. We can thus generate multiple batches of the form $\{(\bs{x}^{(m)}, \bs{\mu}^{(m)})\}_{m=1}^M$ and pass them directly through an invertible network. Since the ground-truth posterior is Gaussian, we can compute the KL divergence as a measure of mismatch between the true and approximate posteriors in closed form.

We run three experiments with $D \in  \{5, 50, 500\}$ where the size of the ACB blocks was doubled for each successive $D$. To asses results, we compute the $R^{2}$ and NRMSE between approximate and true means as well as the KL divergence between approximate and true distributions on 100 test datasets. To compute the approximate covariance matrix, we draw 5000 samples from the approximate posteriors for $D=5$ and $D=50$ and 50000 samples for $D=500$. 

The KL divergence for the 5-D and 50-D MVNs reached essentially 0 after 2-3 epochs of 1000 iterations indicating that this is an easy problem for BayesFlow, and almost perfect recovery of the true posteriors is possible. 
The KL divergence for the 500-D MVN model reached 0.37 after 50 epochs, which represents a negligible increase in entropy relative to the true posterior ($0.05\%$ nats) and
indicates decent approximation in light of the high dimensionality of the problem.



\begin{figure*}
\centering
\begin{subfigure}{.99\textwidth}
\includegraphics[width=\textwidth]{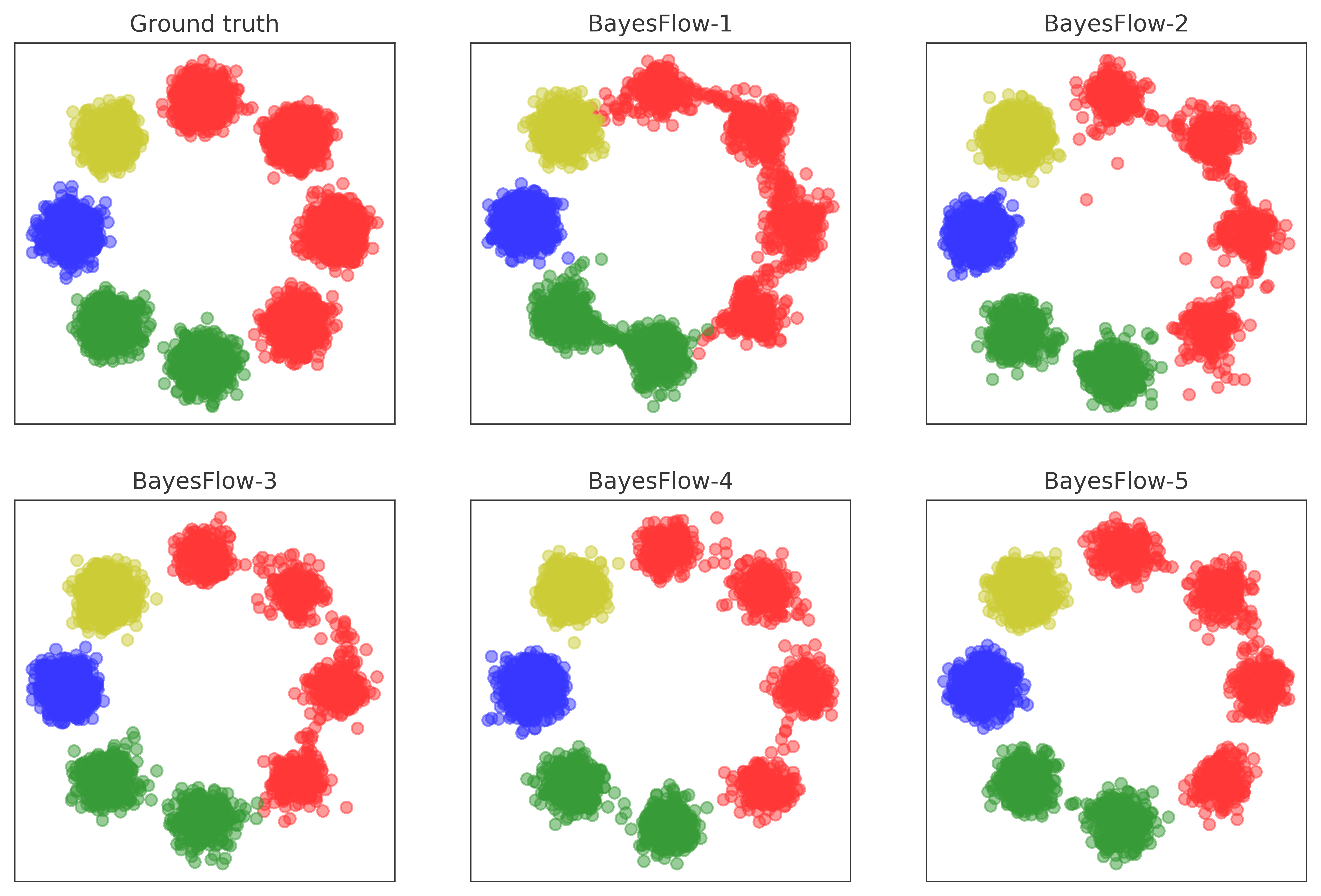}
\end{subfigure}
\caption[short]{Results on the GMM toy example with colors indicating cluster assignments. Approximation of the multimodal posterior become closer to the ground truth distribution with increasing depth (number of ACBs) of the conditional invertible network.} \label{fig:Fig.3}
\end{figure*}

\subsection{Multimodal Posterior - Gaussian Mixture Model}
In order to test whether the BayesFlow method can recover multimodal posteriors, we apply it to a generative Gaussian mixture model (GMM). Multimodal posteriors arise in practice, for example, when forward models are defined as mixtures between different processes, or when models exhibit large multivariate trade-offs in their parameter space (e.g., there are multiple separate regions of posterior density with plausible parameter values). Therefore, it is important to show that our method is able to capture such behavior and does not suffer from mode collapse.

Following \cite{ardizzone2018analyzing}, we construct a scenario in which the observed data $\bs{x}$ is a one-hot encoded vector representing one of the \textit{hard} labels \textit{red}, \textit{green}, \textit{blue}, or \textit{yellow} (i.e., a single observation, thus $N=1$). 
The parameters $\bs{\theta} = (\theta_{1}, \theta_{2})$ are the 2D coordinates
\footnote{Note that this is not the typical GMM setup, as we construct the example such that the mixture assignments (labels) are observed and the data coordinates are the latent parameters.}
of points drawn from a mixture of eight Gaussian clusters with centers distributed around the origin in a clockwise manner and unit variance (see \autoref{fig:Fig.3}, upper left). 
The first four clusters are assigned the label \textit{red}, the next two the label \textit{green}, and the remaining two the labels \textit{blue} and \textit{yellow}. The posterior $p(\bs{\theta}\,|\,\bs{x})$ is composed of the clusters indexed by the corresponding label.
We perform the experiment multiple times by increasing the depth of the BayesFlow starting from 1 ACB block up to 5 ACB blocks. In this way, we can investigate the effects of cINN depth on the quality of the approximate multimodal posteriors. We train each BayesFlow for 50 epochs and draw 8000 samples from the approximate posteriors obtained by the trained models.

Results for all BayesFlows are depicted in \autoref{fig:Fig.3}. We observe that approximations profit from having a deeper cINN chain, with cluster separation becoming clearer when using more ACBs. This confirms that our method is capable of recovering multimodal posteriors. 

\subsection{Stochastic Time-Series Model - The Ricker Model}

In the following, we estimate the parameters of a well-known discrete stochastic population dynamics model \cite{wood2010statistical}. With this example, we are pursuing several goals: First, we want to demonstrate that the BayesFlow method is able to accurately recover the parameters of an actual model with intractable likelihood by learning summary statistics from raw data. Second, we show that BayesFlow can deal adequately with parameters that are completely unrelated to the data by reducing estimates to the corresponding parameters' prior. Third, we compare the global performance of the BayesFlow method to that of related methods capable of amortized likelihood-free inference. Finally, we demonstrate the desired posterior contraction and improvement in estimation with increasing number of observations.

Discrete population dynamics models describe how the number of individuals in a population changes over discrete units of time \cite{wood2010statistical}. In particular, the Ricker model describes the number of individuals $x_{t}$ in generation $t$ as a function of the expected number of individuals in the previous generation by the following non-linear equations:
\begin{align}
x_{t} &\sim Pois(\rho N_{t}) \\
\xi_{t} &\sim \mathcal{N}(0, \sigma^{2}) \\
N_{t+1} &= rN_{t}e^{-N_{t} + \xi_{t}} 
\end{align}
for $t = 1,...,T$ where $N_{t}$ is the expected number of individuals at time $t$, $r$ is the growth rate, $\rho$ is a scaling parameter and $\xi_{t}$ is random Gaussian noise. The likelihood function for the Ricker model is not available in closed form, and the model is known to exhibit chaotic behavior \cite{mestdagh2019prepaid}. Thus, it is a suitable candidate for likelihood-free inference.  The parameter estimation task consists of recovering $\bs{\theta}=(\rho,r,\sigma)$ from the observed one-dimensional time-series data $\bs{x}_{1:T}$ where each $x_{t} \in \mathbb{N}$. 

\begin{figure*}
\centering
\begin{subfigure}{.52\textwidth}
	\includegraphics[width=\textwidth]{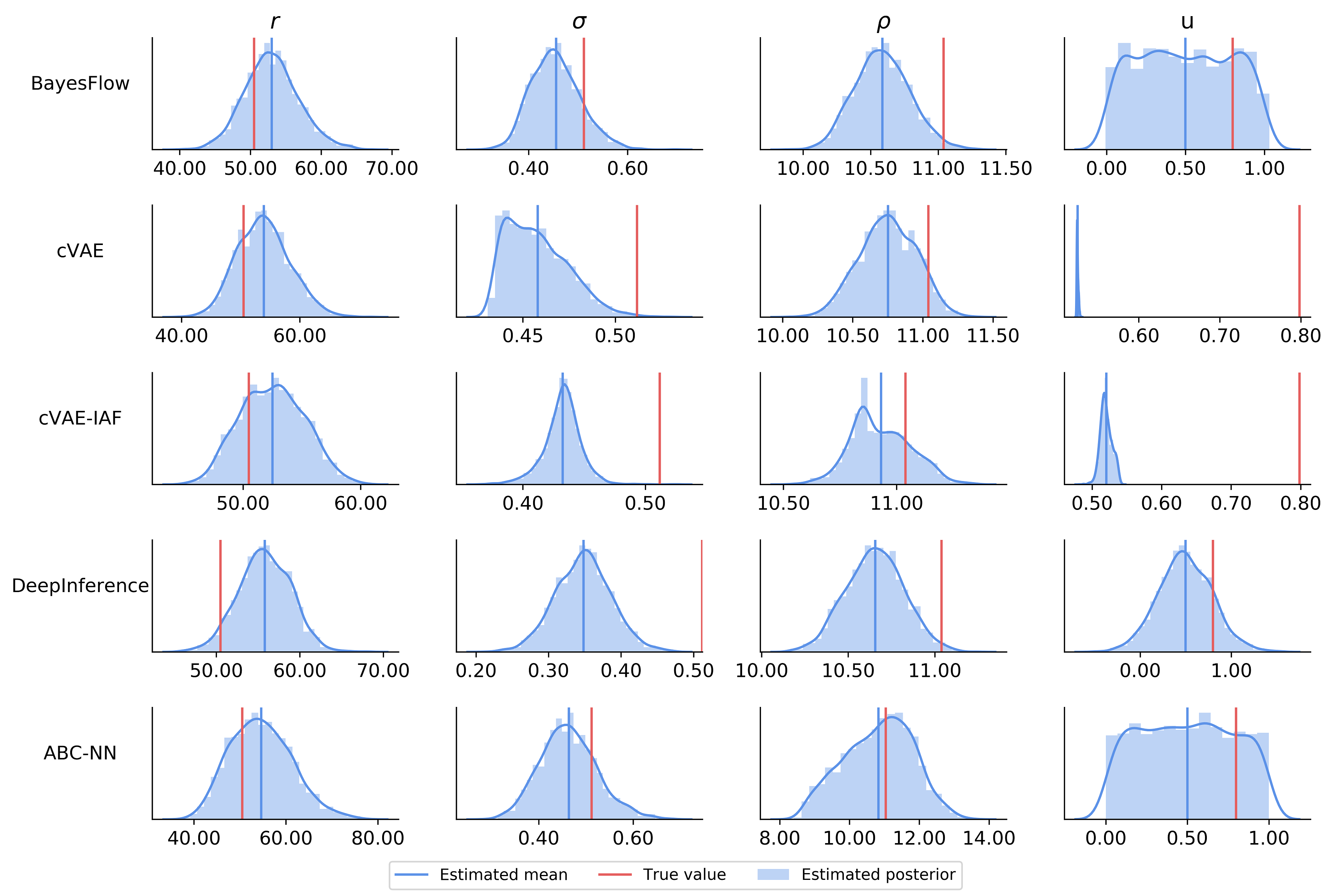}
    \caption{Full posteriors from all methods for an example Ricker dataset.}
    \label{fig:Fig.4a}
\end{subfigure}
\begin{subfigure}{.47\textwidth}
	\begin{subfigure}{.95\textwidth}
		\includegraphics[width=\textwidth]{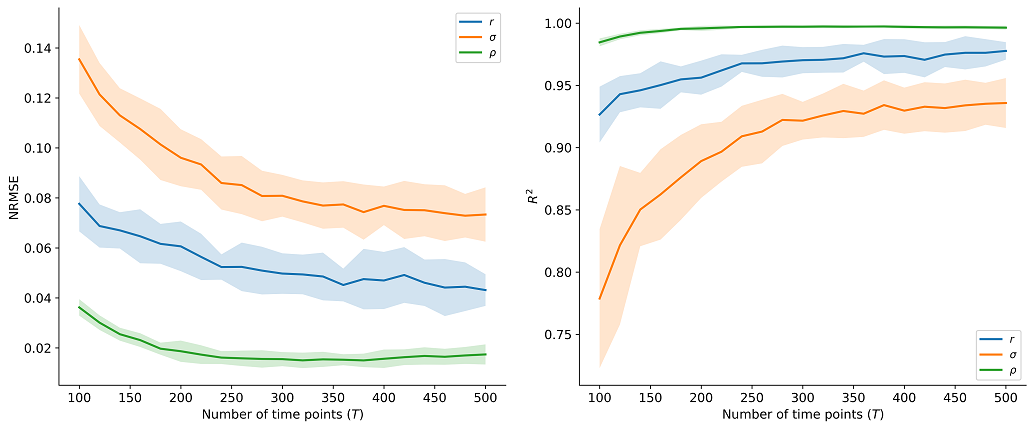}
    	\caption{Performance over all $T$s}
    \label{fig:Fig.4b}
	\end{subfigure}
	\begin{subfigure}{.95\textwidth}
		\includegraphics[width=\textwidth]{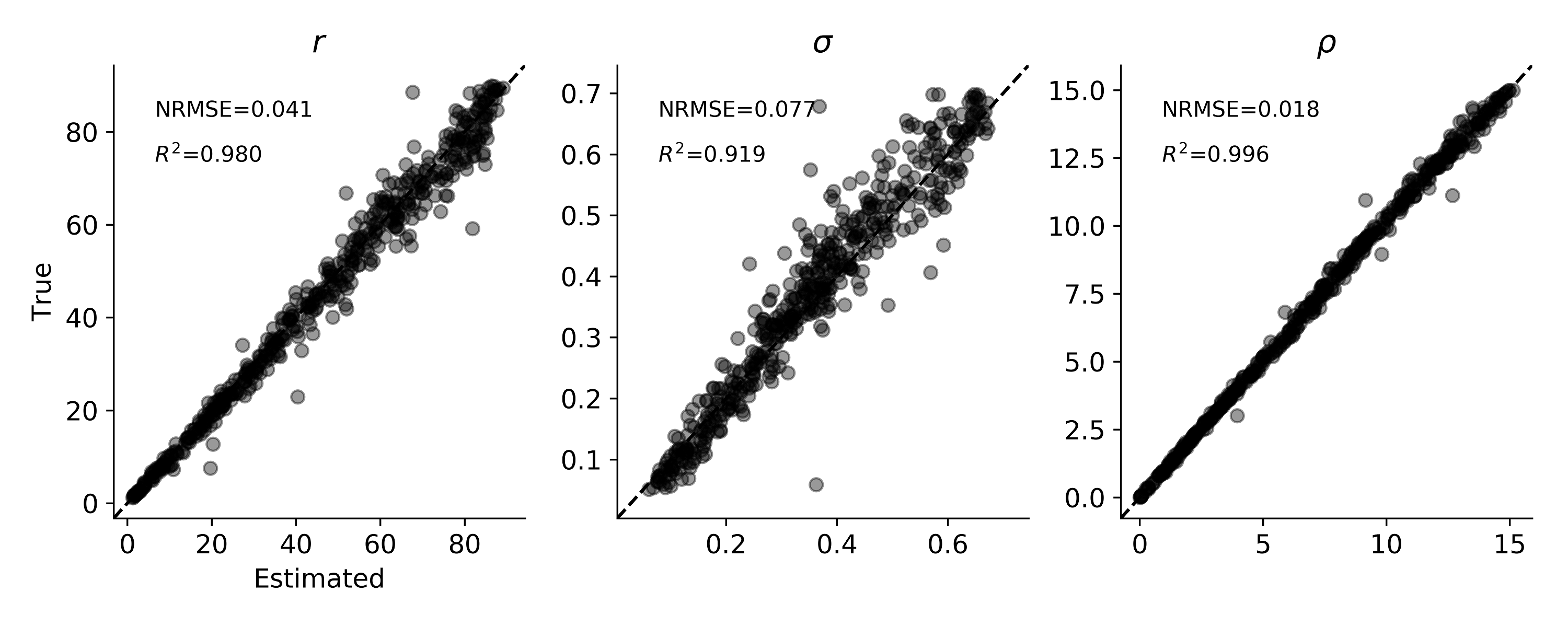}
    	\caption{Parameter recovery ($T=500$)}
    \label{fig:Fig.4c}
	\end{subfigure}
\end{subfigure}
\begin{subfigure}{.95\textwidth}
	\includegraphics[width=\textwidth]{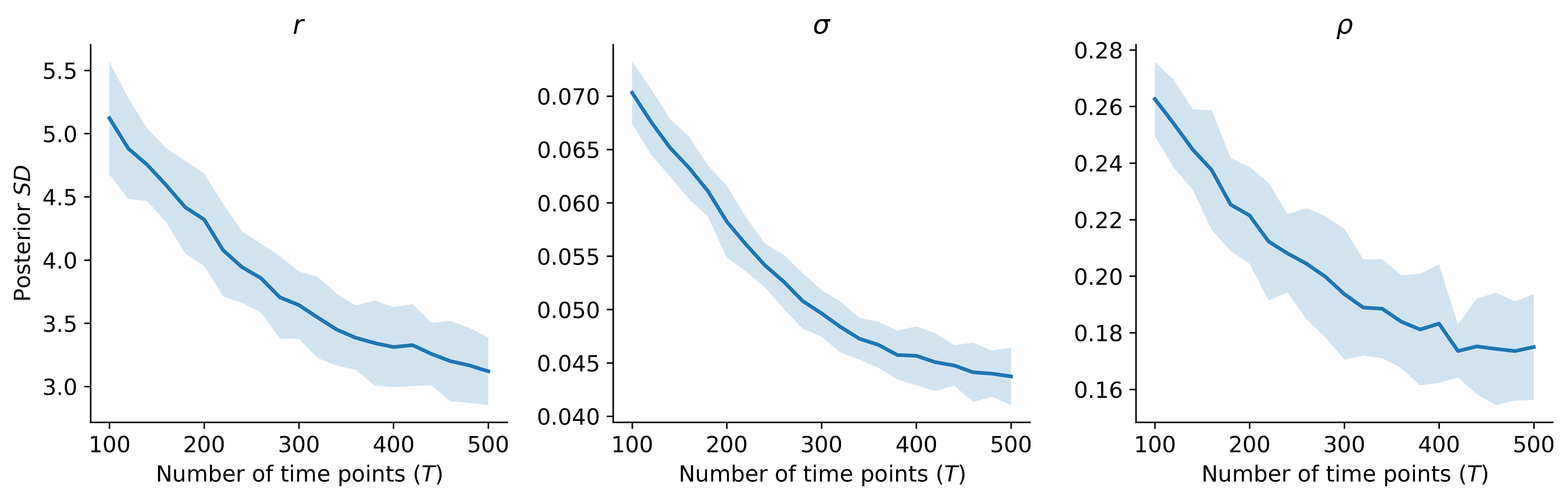}
	\caption{Posterior contraction with increasing $T$}
\label{fig:Fig.4d}
\end{subfigure}
\caption[short]{Results on the Ricker model. \textbf{(a)} Approximate posteriors obtained by all implemented methods on a single Ricker dataset. Note that only BayesFlow and ABC-NN are able to approximate the uniform posterior of $u$; \textbf{(b)} NRMSE and $R^{2}$ performance metrics over all $T$s obtained by the BayesFlow method. We observe that parameter estimation remains good over all $T$s, and becomes progressively better as more data is available (shaded regions indicate bootstrap 95\% CIs); \textbf{(c)} Parameter recovery with BayesFlow for the maximum number of generations used during training ($T=500$); \textbf{(d)} Posterior contraction in terms of posterior standard deviation for each parameter across increasing number of available generations (shaded regions indicate bootstrap 95\% CIs).} \label{fig:Fig.4}
\end{figure*}

What if the data does not contain any information about a particular parameter? In this case, any good estimation method should detect this, and return the prior of the particular parameter. To test this, we append a random uniform variable $u \sim \mathcal{U}(0, 1)$ to the parameter vector $\bs{\theta}$ and train BayesFlow with this additional dummy parameter. We expect that the networks ignore this dummy parameter, that is, we assume that the estimated posterior of $u$ resembles the uniform prior. 

We compare the performance of BayesFlow to the following recent methods capable of amortized likelihood-free inference: conditional variational autoencoder  (cVAE) \cite{mishra2018generative}, cVAE with autoregressive flow (cVAE-IAF) \cite{kingma2016improved}, \textit{DeepInference} with heteroscedastic loss \cite{radev2019towards}, approximate Bayesian computation with an LSTM neural network for learning informative summary statistics (ABC-NN) \cite{jiang2017learning} and quantile random forest (ABC-RF) \cite{raynal2018abc}. For training the models, we simulate time-series from the Ricker model with varying lengths. The number of time points $T$ is drawn from a uniform distribution $T \sim \mathcal{U}(100, 500)$ at each training iteration. 

All neural network methods were trained for $100$ epochs with $1000$ iterations each on simulated data from the Ricker model. The ABC-RF method was fitted on a reference table with $200\ 000$ datasets, since the method does not allow for online learning and increasing the reference table did not seem to improve performance. In order to avoid using hand-crafted summary statistics for the ABC-RF method, we input summary vectors obtained by applying the summary network trained jointly with the cINN. Thus, the ABC-RF method has the advantage of using maximally informative statistics as input. We validate the performance of all methods on an independent test set of 500 datasets generated with $T = 500$. We report performance metrics for each method and each parameter in \autoref{table:Table 1}.

Parameters $r$ and $\rho$ seem to be well recoverable by all methods considered here. The $\sigma$ parameter turns out to be harder to estimate, with BayesFlow and the ABC-NN method performing best. Further, BayesFlow performs very well across all parameters and metrics. Importantly, the calibration error $Err_{cal}$ obtained by BayesFlow is always low, indicating that the shape of the approximate posterior closely matches that of the true posteriors. Variational methods (cVAE, cVAE-IAF) experience some problems recovering the posterior of $\sigma$. The ABC-NN and ABC-RF methods seem to recover point estimates with high accuracy but the approximate posteriors of the former exhibit relatively high calibration error. The ABC-RF method can only estimate posterior quantiles, so no comparable calibration metric could be computed.

Further results are depicted in \autoref{fig:Fig.4}. Inspecting the full posteriors obtained by all methods on an example test dataset, we note that only BayesFlow and the ABC-NN methods are able to recover the uninformative posterior distribution of the dummy noise variable $u$ (\autoref{fig:Fig.4a}). Moreover, the importance of a Bayesian treatment of the Ricker model becomes clear when looking at the posteriors of $\sigma$. On most test datasets, the posterior density spreads over the entire prior range (high posterior variance) indicating large uncertainty in the obtained estimates. Moreover, the shapes of the marginal parameter posteriors vary widely across validation datasets, which highlights the importance of avoiding \textit{ad hoc} restrictions on allowed posterior shapes (see \autoref{fig:Fig.S6} for examples). We also observe that parameter estimation with BayesFlow becomes increasingly accurate when more time points are available (\autoref{fig:Fig.4b}). Parameter recovery is especially good with the maximum number of time points (see \autoref{fig:Fig.4c}). Finally, (\autoref{fig:Fig.4d}) reveals a notable posterior contraction across increasing number of time points available to the summary network. 

\begin{table*}
\centering
\caption{Performance results on the Ricker model across all estimation methods}
\resizebox{\textwidth}{!}{
\begin{tabular}{llllllll}
\toprule
   &     &            BayesFlow &            cVAE &        cVAE-IAF &           DeepInference &          ABC-NN &            ABC-RF\\
\midrule
$Err_{cal}$ & $r$ &  0.017 $\pm$ 0.007 &  \textbf{0.014} $\pm$ \textbf{0.007} &  0.058 $\pm$ 0.017 &  0.122 $\pm$ 0.016 &  0.164 $\pm$ 0.015 & - \\
   & $\sigma$ &  \textbf{0.013} $\pm$ \textbf{0.007} &  0.419 $\pm$ 0.011 &  0.382 $\pm$ 0.013 &  0.184 $\pm$ 0.021 &  0.119 $\pm$ 0.014 & - \\
   & $\rho$ &  \textbf{0.084} $\pm$ \textbf{0.018} &  0.121 $\pm$ 0.017 &  0.188 $\pm$ 0.018 &  0.111 $\pm$ 0.019 &  0.283 $\pm$ 0.012 & - \\
$NRMSE$ & $r$ &  \textbf{0.041} $\pm$ \textbf{0.002} &  0.047 $\pm$ 0.004 &  0.047 $\pm$ 0.006 &  0.052 $\pm$ 0.003 &  0.053 $\pm$ 0.003 & 0.044 $\pm$ 0.004 \\
   & $\sigma$ &  \textbf{0.077} $\pm$ \textbf{0.005} &  0.137 $\pm$ 0.004 &  0.124 $\pm$ 0.006 &  0.108 $\pm$ 0.004 &  \textbf{0.077} $\pm$ \textbf{0.004} & 0.081 $\pm$ 0.005 \\
   & $\rho$ &  0.018 $\pm$ 0.001 &  \textbf{0.016} $\pm$ \textbf{0.002} &  0.019 $\pm$ 0.002 &  0.019 $\pm$ 0.002 &  0.033 $\pm$ 0.002 & 0.021 $\pm$ 0.001 \\
$R^{2}$ & $r$ &  \textbf{0.980} $\pm$ \textbf{0.003} &  0.973 $\pm$ 0.005 &  0.973 $\pm$ 0.007 &  0.968 $\pm$ 0.005 &  0.966 $\pm$ 0.004 & 0.977 $\pm$ 0.004  \\
   & $\sigma$ &  \textbf{0.919} $\pm$ \textbf{0.011} &  0.745 $\pm$ 0.020 &  0.792 $\pm$ 0.020 &  0.841 $\pm$ 0.014 &  \textbf{0.919} $\pm$ \textbf{0.010} & 0.912 $\pm$ 0.011 \\
   & $\rho$ &  0.996 $\pm$ 0.001 &  \textbf{0.997} $\pm$ \textbf{0.001} &  0.996 $\pm$ 0.001 &  0.996 $\pm$ 0.001 &  0.986 $\pm$ 0.002 & 0.994 $\pm$ 0.001 \\
   
$Err_{sim}$ & - &  \textbf{0.038} $\pm$ \textbf{0.001} &  0.041 $\pm$ 0.001 &  0.042 $\pm$ 0.001 &  0.041 $\pm$ 0.001 &  0.048 $\pm$ 0.002 & 0.041 $\pm$ 0.002 \\
\bottomrule
\end{tabular}}
\begin{tablenotes}
      \small
      \item Note: For each parameter, bootstrapped means ($\pm 1$ standard error) of different performance metrics are displayed for all tested methods. 
For each metric and each parameter, the best performance across methods is printed in bold font.
\end{tablenotes}
\label{table:Table 1}
\end{table*}

\subsection{A Model of Perceptual Decision Making - The Lévy-Flight Model}

In the following, we estimate the parameters of a stochastic differential equation model of human decision making. We perform the first Bayesian treatment of the recently proposed Lévy-Flight Model (LFM), as its intractability has so far rendered traditional non-amortized Bayesian inference methods prohibitively slow \cite{voss2019sequential}. 

With this example, we first want to show empirically that BayesFlow is able to deal with $i.i.d.$ datasets of variable size arising from $N$ independent runs of a complex stochastic simulator. For this, we inspect global performance of BayesFlow over a wide range of dataset sizes. Additionally, we want to show the advantage of amortized inference compared to case-based inference in terms of efficiency and recovery. For this, we apply BayesFlow along with four other recent methods for likelihood-free inference to a single dataset and show that in some cases the speed advantage of amortized inference becomes noticeable even after as few as 5 datasets. Crucially, researchers often fit the same models to different datasets, so if a pre-trained model exists, it would present a huge advantage in terms of efficiency and productivity.

We focus on the family of evidence accumulator models (EAMs) which describe human decision making by a set of neurocognitively motivated parameters \cite{ratcliff2008diffusion}. EAMs are most often applied to choice reaction times (RT) data to obtain an estimate of the underlying processes governing categorization and (perceptual) decision making. In its most general formulation, the forward model of EAMs takes the form of a stochastic ordinary differential equation (ODE):
\begin{align} 
dx = vdt + \xi\sqrt{dt}  
\end{align}
where $dx$ denotes a change in activation of an accumulator, $v$ denotes the average speed of information accumulation (often termed the drift rate), and $\xi$ represents a stochastic additive component, usually modeled as coming from a Gaussian distribution centered around $0$: $\xi \sim \mathcal{N}(0, c^{2})$.

EAMs are particularly amenable for likelihood-free inference, since the likelihood of most interesting members of this model family turn out to be intractable \cite{miletic2017parameter}. This intractability has precluded many interesting applications and empirically driven model refinements. Here, we apply BayesFlow to estimate the parameters of the recently proposed Lévy-Flight Model (LFM) \cite{voss2019sequential}. The LFM assumes an $\alpha$-stable noise distribution of the evidence accumulation process which allows to model discontinuities in the decision process. However, the inclusion of $\alpha$-stable noise (instead of the typically assumed Gaussian noise) leads to a model with intractable likelihood:
\begin{align}
dx &= vdt + \xi dt^{1/\alpha}  \\
\xi &\sim AlphaStable(\alpha,0,1,0) 
\end{align}
where $\alpha$ controls the probability of outliers in the noise distribution. The LFM has three additional parameters: the threshold $a$ determining the amount of evidence needed for the termination of a decision process; a relative starting point, $zr$, determining the amount of starting evidence available to the accumulator before the actual decision alternatives are presented; and an additive non-decision time $t_{0}$. 

\begin{figure*}
\centering
\begin{subfigure}{.95\textwidth}
    \includegraphics[width=\textwidth]{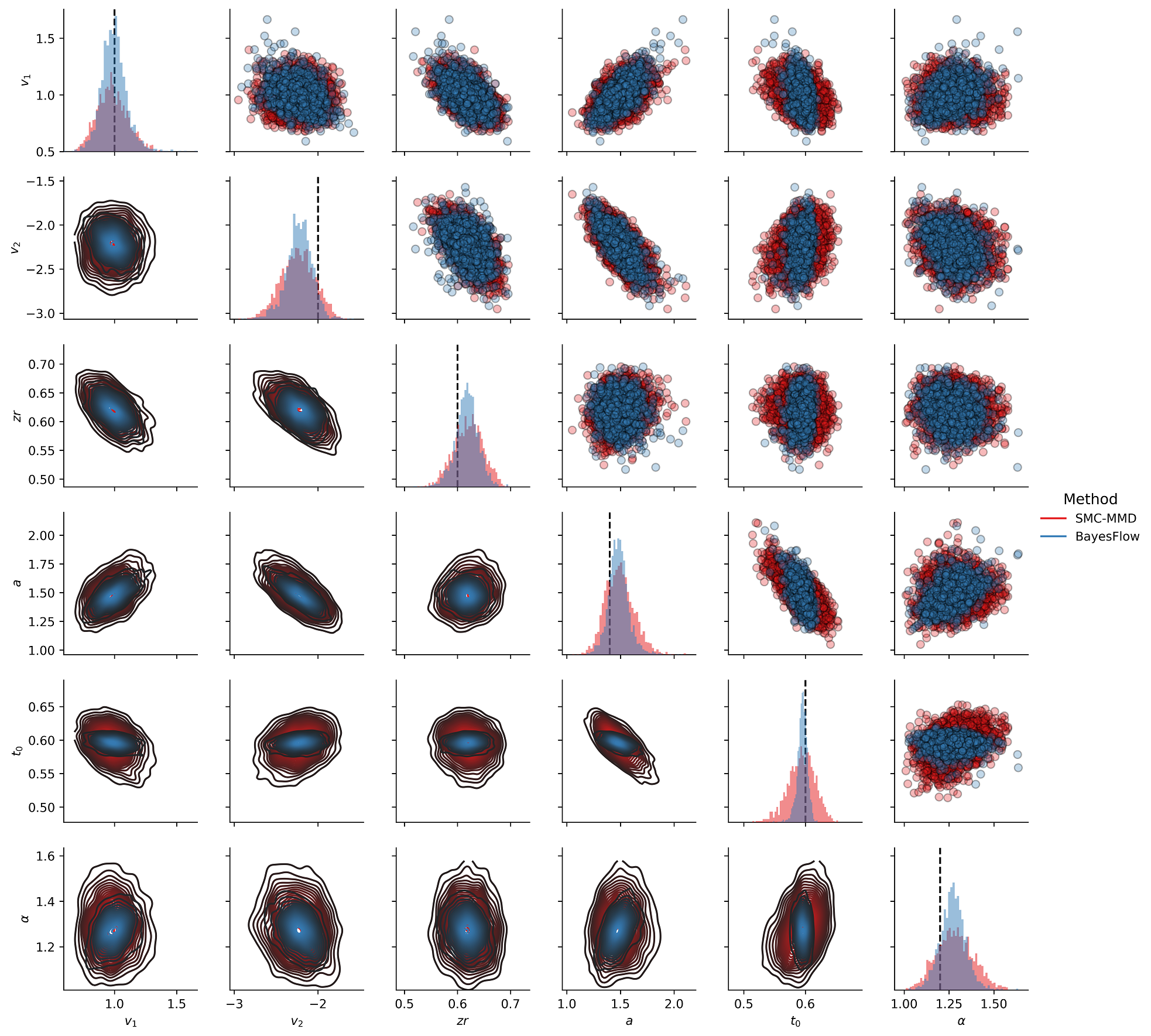}
    \caption{Joint posteriors from BayesFlow and SMC-MMD}
    \label{fig:Fig.5a}
\end{subfigure}
\begin{subfigure}{.95\textwidth}
    \includegraphics[width=\textwidth]{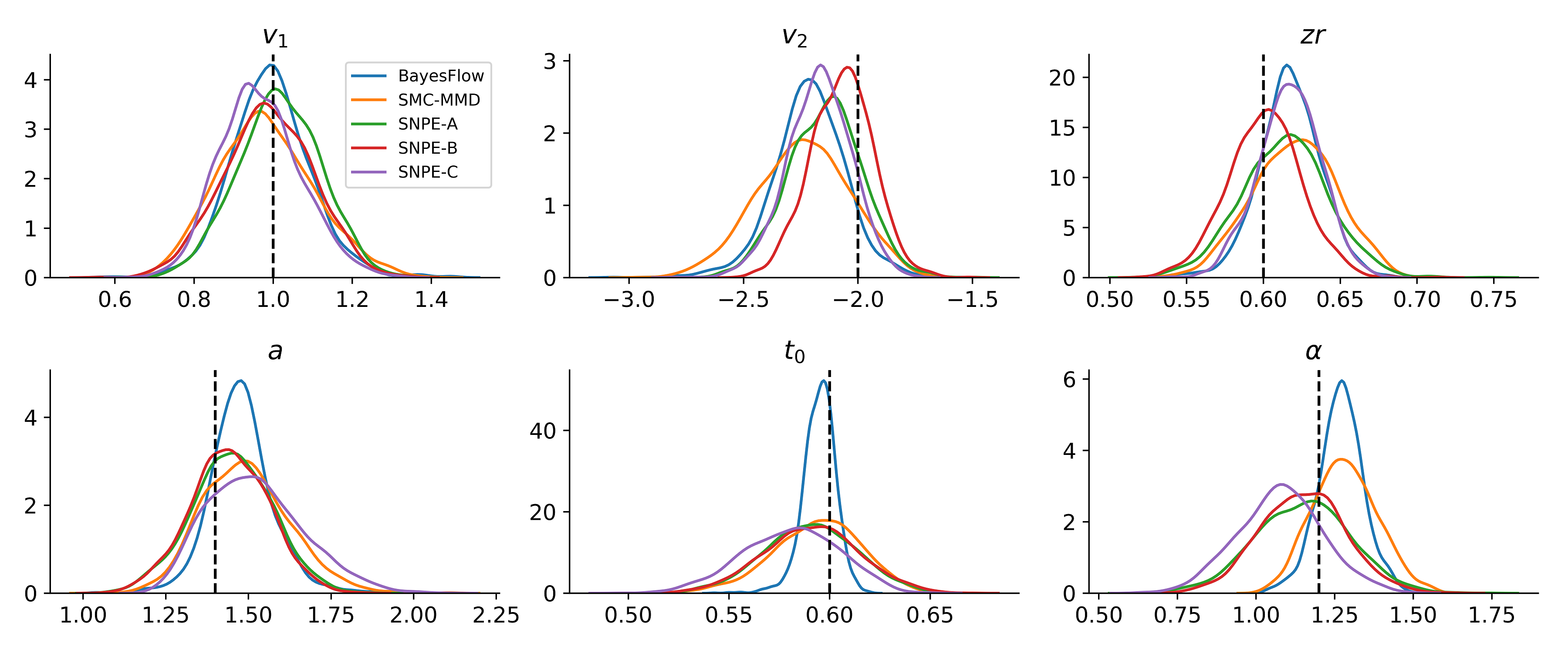}
    \caption{Marginal posteriors from all methods}
    \label{fig:Fig.5b}
\end{subfigure}
\caption[short]{Comparison results on the LFM model. \textbf{(a)} Marginal and bivariate posteriors obtained by BayesFlow and SMC-MMD on the single validation dataset. We observe markedly better sharpness in the BayesFlow posteriors; \textbf{(b)} Marginal posteriors obtained from all methods under comparison.} \label{fig:Fig.5}
\end{figure*}

During training of the networks, we simulate response times data from two experimental conditions with two different drift rates, since such a design is often encountered in psychological research. The parameter estimation task is thus to recover the parameters $\bs{\theta} = (v_{0}, v_{1}, a, t_{0}, zr, \alpha)$ from two-dimensional \textit{i.i.d.} RT data $\bs{x}_{1:N}$ where each $\bs{x}_{i} \in \mathbb{R}^{2}$ represents RTs obtained in the two conditions. The number of trials is drawn from a uniform distribution $N \sim \mathcal{U}(100, 1000)$ at each training iteration. Training the networks took a little less than a day with the online learning approach. Inference on $1000$ datasets with $2000$ posterior samples per parameter took approximately $7.39$ seconds. 

In order to investigate whether amortized inference is advantageous for this model, we additionally apply a version of the SMC-ABC algorithm available in the \textit{pyABC} package \cite{klinger2018pyabc} to a single dataset with $N = 500$. Since no sufficient summary statistics are available for EAM data, we apply the maximum mean discrepancy (MMD) metric as a distance between the full raw empirical RT distributions, in order to prevent information loss \cite{park2016k2}. Since the MMD is expensive to compute, we use a GPU implementation to ensure that computation of MMD is not a bottleneck for the comparison. In order to achieve good approximation with 2000 samples from the SMC-MMD approximate posterior, we run the algorithm for 20 populations with a final rejection threshold $\epsilon = 0.04$. We also draw $2000$ samples from the approximate posterior obtained by applying our pre-trained BayesFlow networks to the same dataset. 

Along SMC-MMD, we apply three recent methods for neural density estimation, SNPE-A \cite{papamakarios2016fast}, SNPE-B \cite{lueckmann2017flexible}, and SNPE-C (\cite{greenberg2019automatic}, also dubbed APT). Since these methods all depend on summary statistics of the data, we compute the first 6 moments of each empirical response time distribution as well as the fractions of correct/wrong responses. We train each method for a single round with $100$ epochs and $5000$ simulated datasets, in order to keep running time at a minimum. Also, we did not observe improvement in performance when training for more than one round. For each model, we sample $2000$ samples from the approximate joint posterior to align the number of samples with those obtained via SMC-MMD.

\begin{figure}
\centering
\begin{subfigure}{.49\textwidth}
    \includegraphics[width=\textwidth]{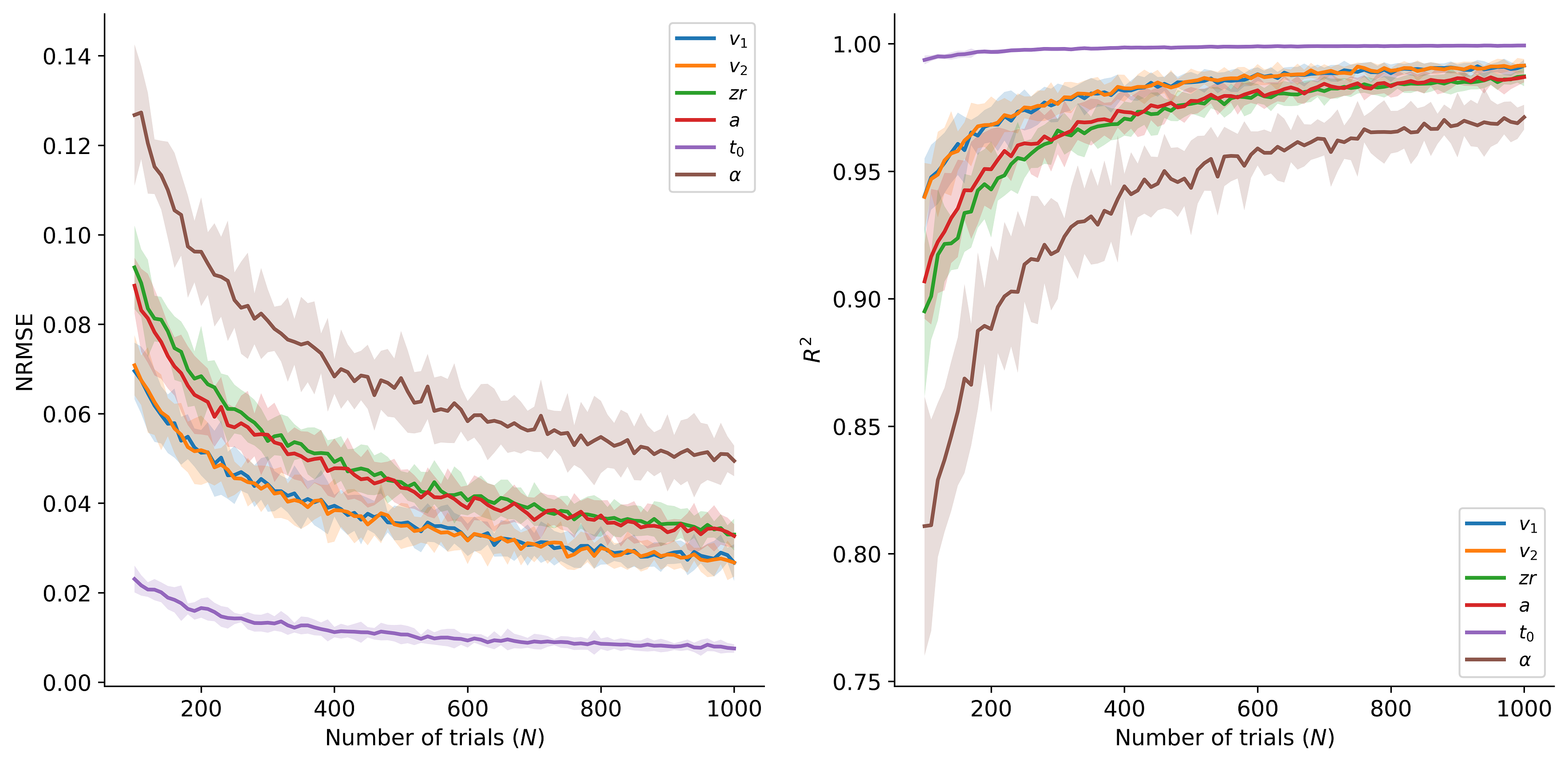}
    \caption{Performance over all trial numbers $N$}
    \label{fig:Fig.6a}
\end{subfigure}
    \begin{subfigure}{.49\textwidth}
    \includegraphics[width=\textwidth]{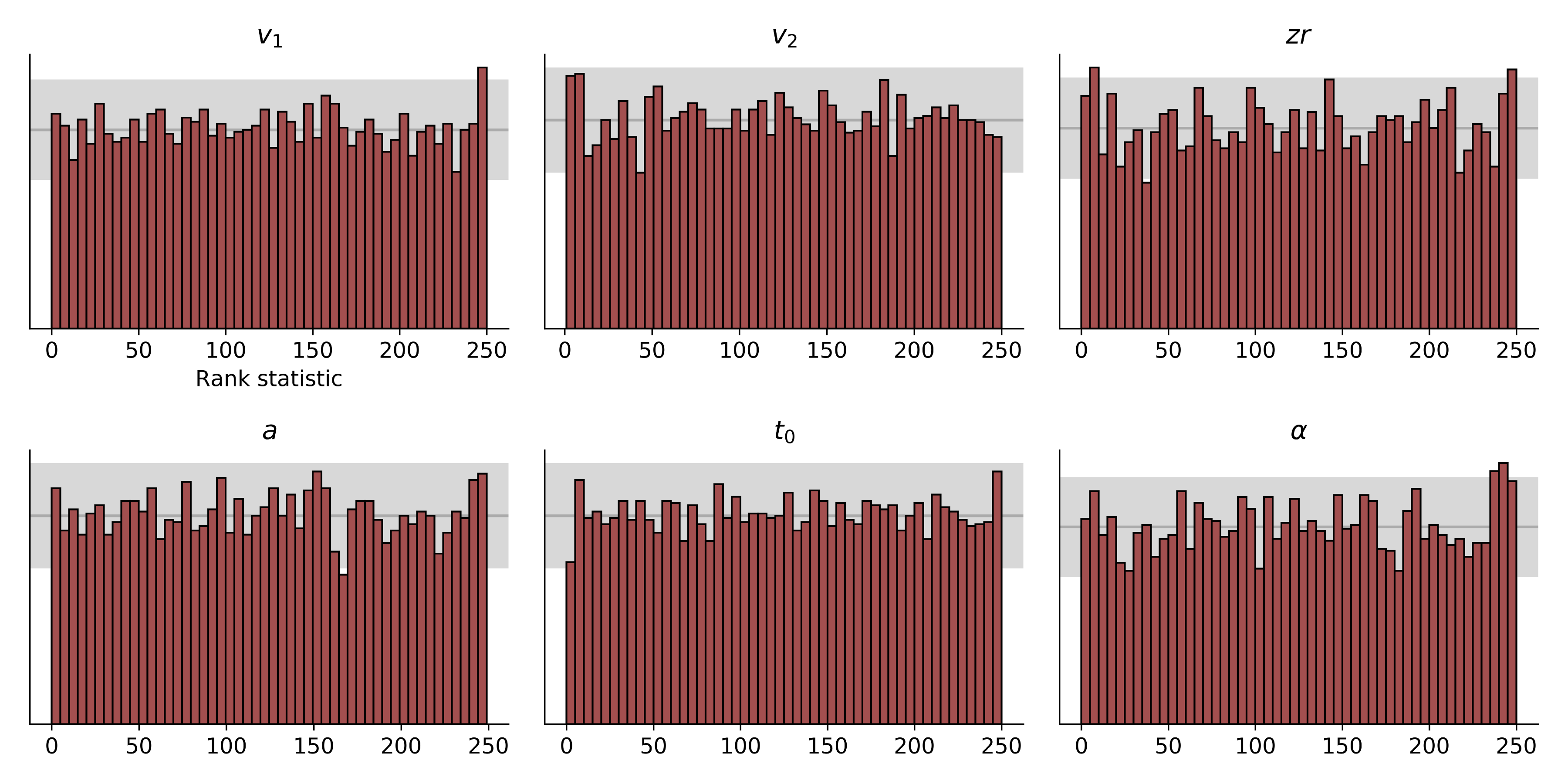}
    \caption{Simulation-based calibration (SBC) at $N=1000$}
    \label{fig:Fig.6b}
\end{subfigure}
\caption[short]{BayesFlow results obtained on the LFM model.} \label{fig:Fig.6}
\end{figure}

The comparison results are depicted in \autoref{fig:Fig.5}. We first focus on the comparison with SMC-MMD on the single dataset. \autoref{fig:Fig.5a} depicts marginal and bivariate posteriors obtained by BayesFlow and SMC-MMD. The approximate posteriors of BayesFlow appear noticeably sharper. Observing the SCB plots (\autoref{fig:Fig.6b}), we can conclude that the approximate posteriors of BayesFlow mirror the sharpness of the true posterior, since otherwise the SCB plots would show marked deviations from uniformity. Further, \autoref{fig:Fig.5b} depicts the marginal posteriors obtained from the application of each method. Noticeably, performance and sharpness varies across the methods and parameters, with all methods yielding good point-estimate recovery via posterior means in terms of the NRMSE and $R^2$ metrics. 

\begin{table}
\centering
\caption{Speed of inference and break-even for amortized inference for the LFM model}
\resizebox{.99\textwidth}{!}{
\begin{tabular}{lllll}
\toprule
   &  Upfront Training & Inference (1 dataset) & Inference (500 datasets)  & Break-even after \\
\midrule
BayesFlow & 23.2 h & 60 ms & 3.7 s & - \\
SMC-MMD & - & 5.5 h & 2700 h & 5 datasets \\
SNPE-A & - & 0.65 h & 325 h & 37 datasets \\
SNPE-B & - & 0.65 h & 325 h & 37 datasets \\
SNPE-C & - & 0.35 h & 175 h & 75 datasets \\
\bottomrule
\end{tabular}}
\begin{tablenotes}
      \small
      \item Note: Inference times for 500 datasets as well as the number of datasets for break-even with BayesFlow for the SMC-MMD, SNPE-A, SNPE-B, and SNPE-C methods are extrapolated from the wall-clock running time on a single dataset, so these are approximate quantities.
\end{tablenotes}
\label{table:Table 2}
\end{table}

Importantly, \autoref{table:Table 2} summarizes the advantage of amortized inference for the LFM model in terms of efficiency. For instance, compared to SMC-MMD, the extra effort of learning a global BayesFlow model upfront is worthwhile even after as few as 5 datasets, as inference with SMC-MMD would have taken more than a day to finish. On the other hand, the break-even for SNPE-C/APT occurs after 75 datasets, so in cases where only a few dozens of datasets are considered, case-based inference might be preferable. However, the difficulties in manually finding meaningful and efficiently computable summary statistics may eat up possible savings even in this situation. We acknowledge that our choices in this respect might be sub-optimal, so performance comparisons should be treated with some caution. 

We note, that after a day of training, the pre-trained networks of BayesFlow take less than $5$ seconds to perform inference on $500$ datasets even with maximum number of trials $N=1000$. Using the case-based SMC-MMD algorithm, $500$ inference runs would have taken more than half a year to complete. We also note, that parallelizing separate inference threads across multiple cores or across nodes of a (GPU) computing cluster can dramatically increase the wall-clock speed of the case-based methods considered here. However, the same applies to BayesFlow training, since its most expensive part, the simulation from the forward model, would profit the most from parallel computing. 

The global performance of BayesFlow over all validation datasets and all trial sizes $N$ is depicted in (\autoref{fig:Fig.6}). First, we observe excellent recovery of all LFM parameters with NRMSEs ranging between $0.008$ and $0.048$ and $R^{2}$ between $0.972$ and $0.99$ for the maximum number of trials. Importantly, estimation remains very good across all trial numbers, and improves as more trials become available (\autoref{fig:Fig.6a}). The parameter $\alpha$ appears to be most challenging to estimate, requiring more data for good estimation, whereas the non-decision time parameter $t_{0}$ can be recovered almost perfectly for all trial sizes. Last, the SCB histograms indicate no systematic deviations across the marginal posteriors (\autoref{fig:Fig.6b}).

\subsection{Stochastic Differential Equations - The SIR Epidemiology Model}

With this example, we want to further corroborate the excellent global performance and probabilistic calibration observed for the LFM model on a non-\textit{i.i.d.} stochastic ODE model. For this, we study a compartmental model from epidemiology, whose output comprises variable-sized multidimensional and inter-dependent time-series. It is therefore of interest to investigate how our method performs when applied to data which is the direct output of an ODE simulator.

Compartmental models in epidemiology describe the stochastic dynamics of infectious diseases as they spread over a population of individuals \cite{keeling2011modeling, hethcote2000mathematics}. The parameters of these models encode important characteristics of diseases, such as infection and recovery rates. The stochastic SIR model describes the transition of $N$ individuals between three discrete states -- susceptible ($S$), infected ($I$), and recovered ($R$) -- whose dynamics follow the equations:
\begin{align}
\triangle S &= -\triangle N_{SI}  \\
\triangle I &= \triangle N_{SI} - \triangle N_{IR}  \\
\triangle R &= \triangle N_{IR} \\
\triangle N_{SI} &\sim Binomial(S, 1 - \exp\left(-\beta\frac{I}{N}\triangle t \right))  \\
\triangle N_{IR} &\sim Binomial(I, 1 - \exp\left(-\gamma\triangle t \right)) 
\end{align}
where $S + I + R = N$ give the number of susceptible, infected, and recovered individuals, respectively. The parameter $\beta$ controls the transition rate from being susceptible to infected, and $\gamma$ controls the transition rate from being infected to recovered. The above listed stochastic system has no analytic solution and thus requires numerical simulation methods for recovering parameter values from data. Cast as a parameter estimation task, the challenge is to recover $\bs{\theta} = \{\beta,\gamma\}$ from three dimensional time-series data $\bs{x}_{1:T}$ where each $\bs{x}_{t} \in \mathbb{N}^{3}$ is a triple containing the number of susceptible ($S$), number of infected ($I$), and recovered ($R$) individuals at time $t$. 

\begin{figure*}
\centering
\begin{subfigure}{.49\textwidth}
	\includegraphics[width=\textwidth]{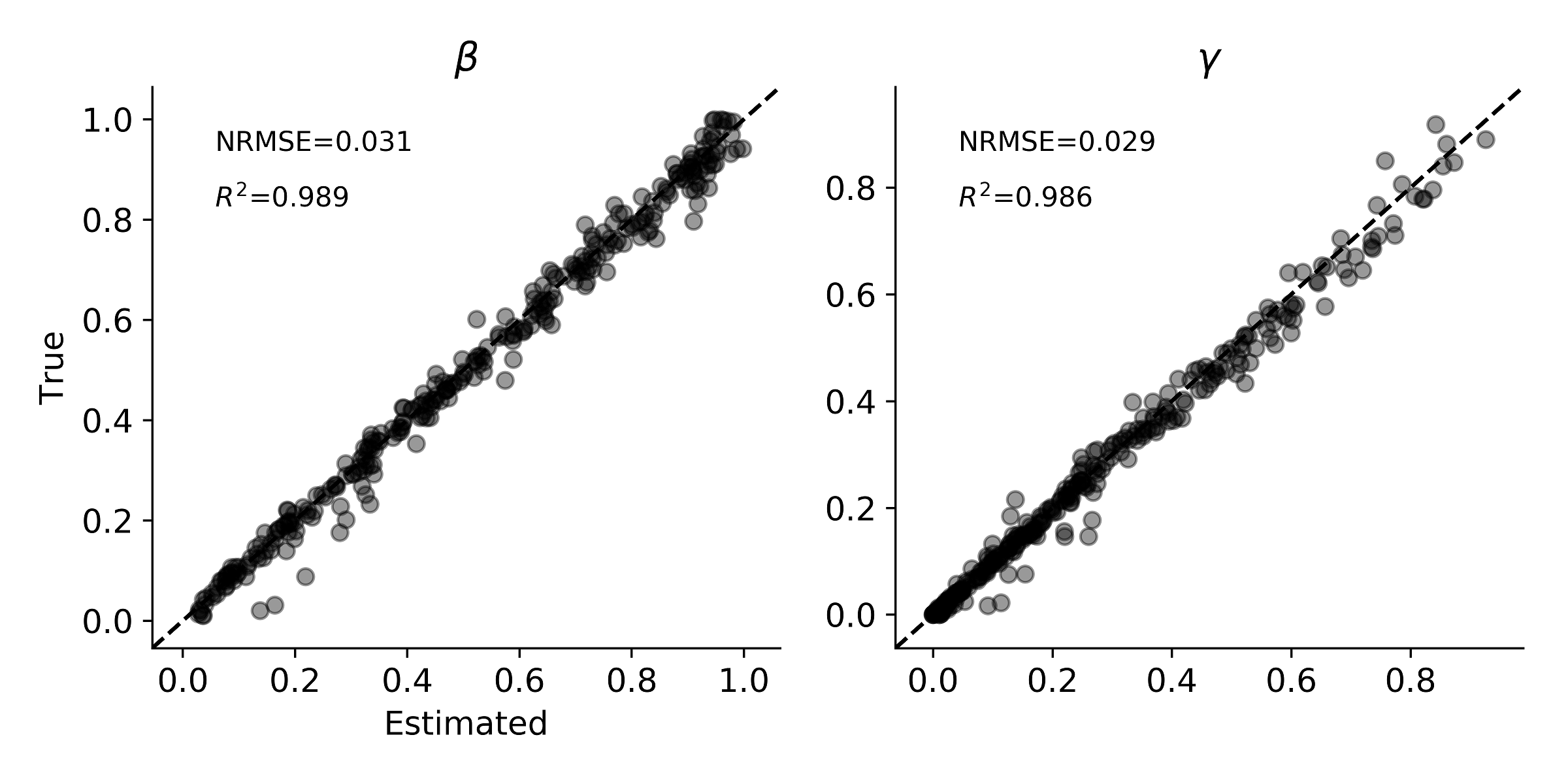}
    \caption{Parameter recovery ($T=500$)}
    \label{fig:Fig.7a}
\end{subfigure}
\begin{subfigure}{.49\textwidth}
	\includegraphics[width=\textwidth]{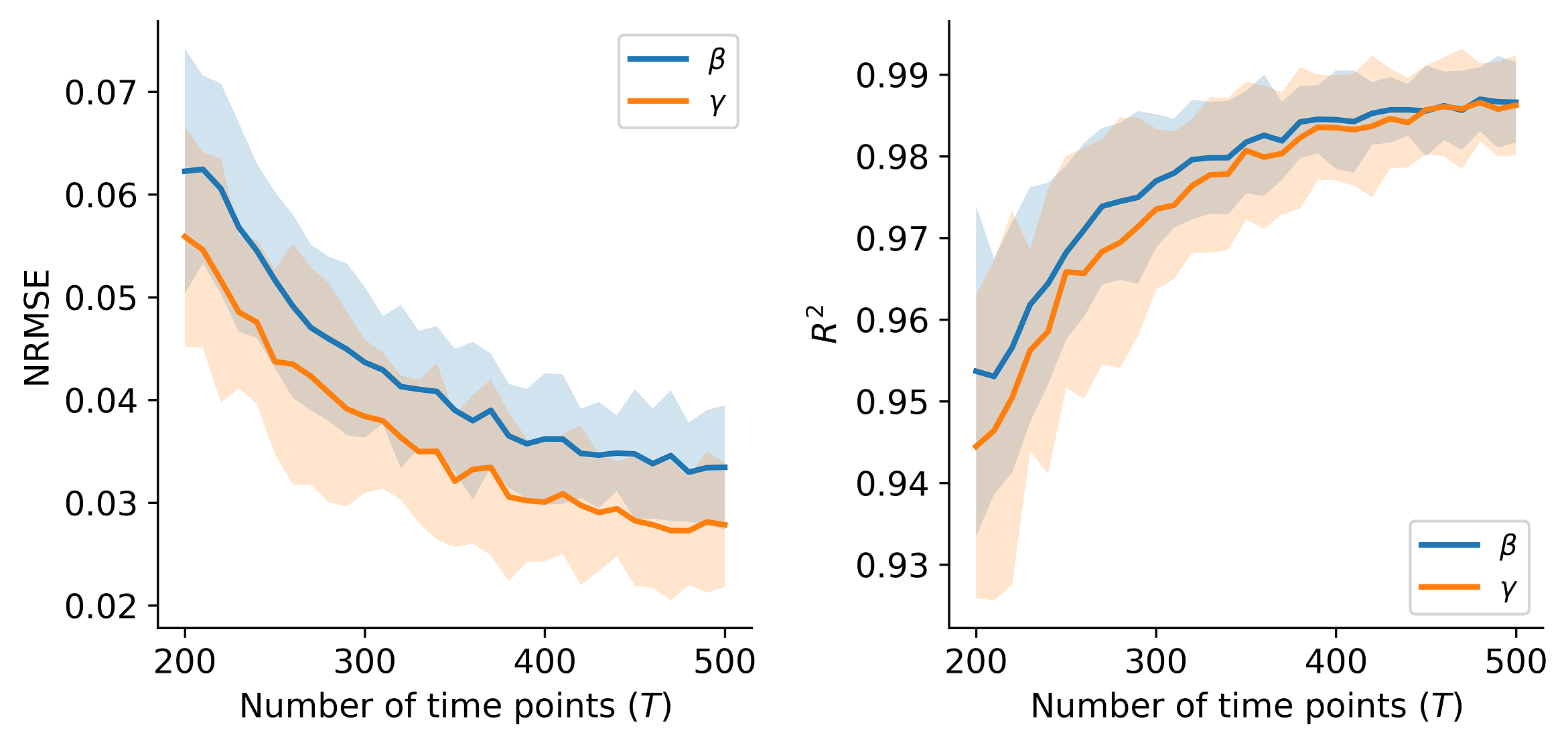}
    \caption{Performance over all $Ts$}
    \label{fig:Fig.7b}
\end{subfigure}
\begin{subfigure}{.49\textwidth}
	\includegraphics[width=\textwidth]{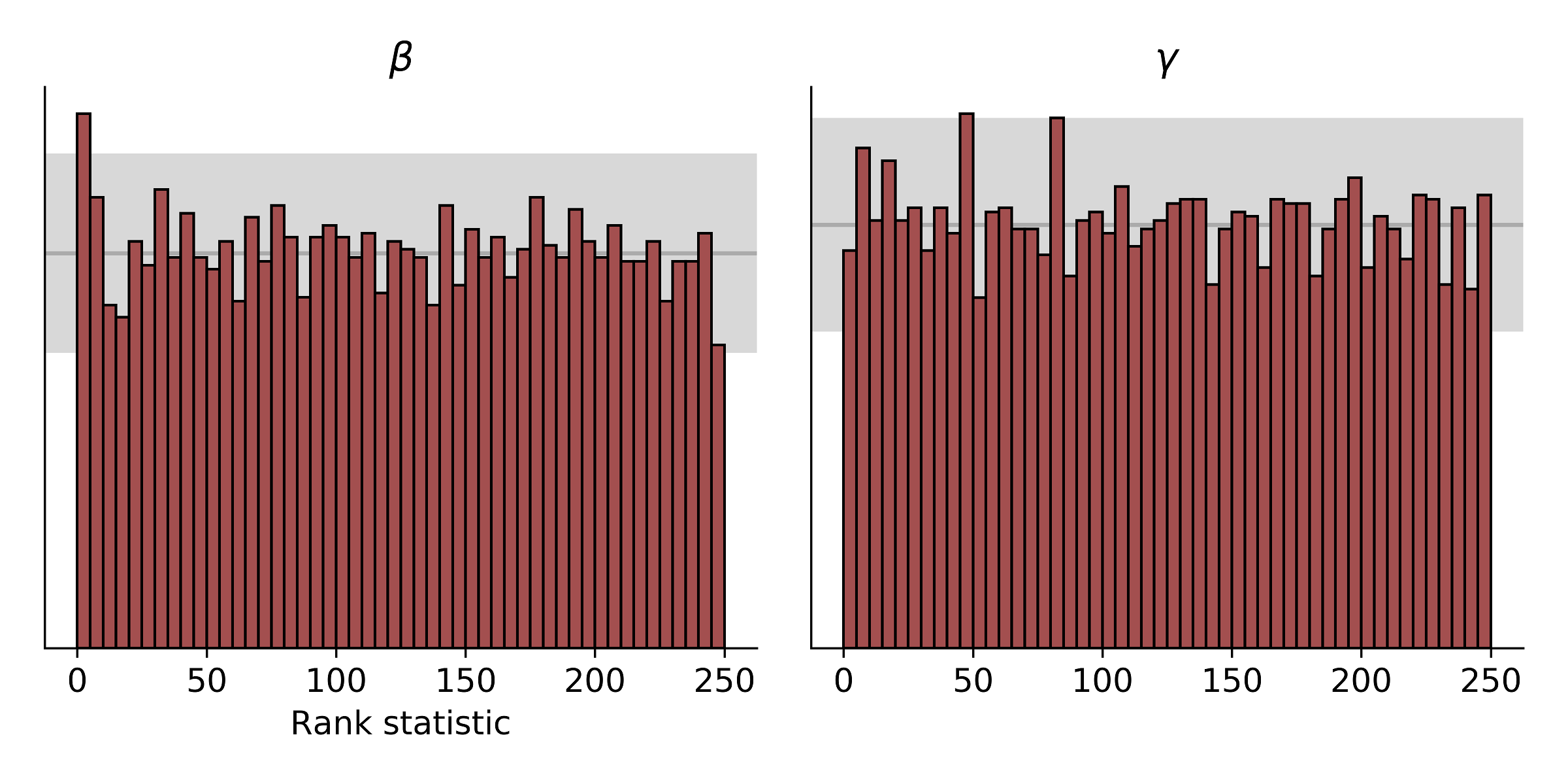}
    \caption{Simulation-based calibration (SBC)}
    \label{fig:Fig.7c}
\end{subfigure}
\begin{subfigure}{.49\textwidth}
	\includegraphics[width=\textwidth]{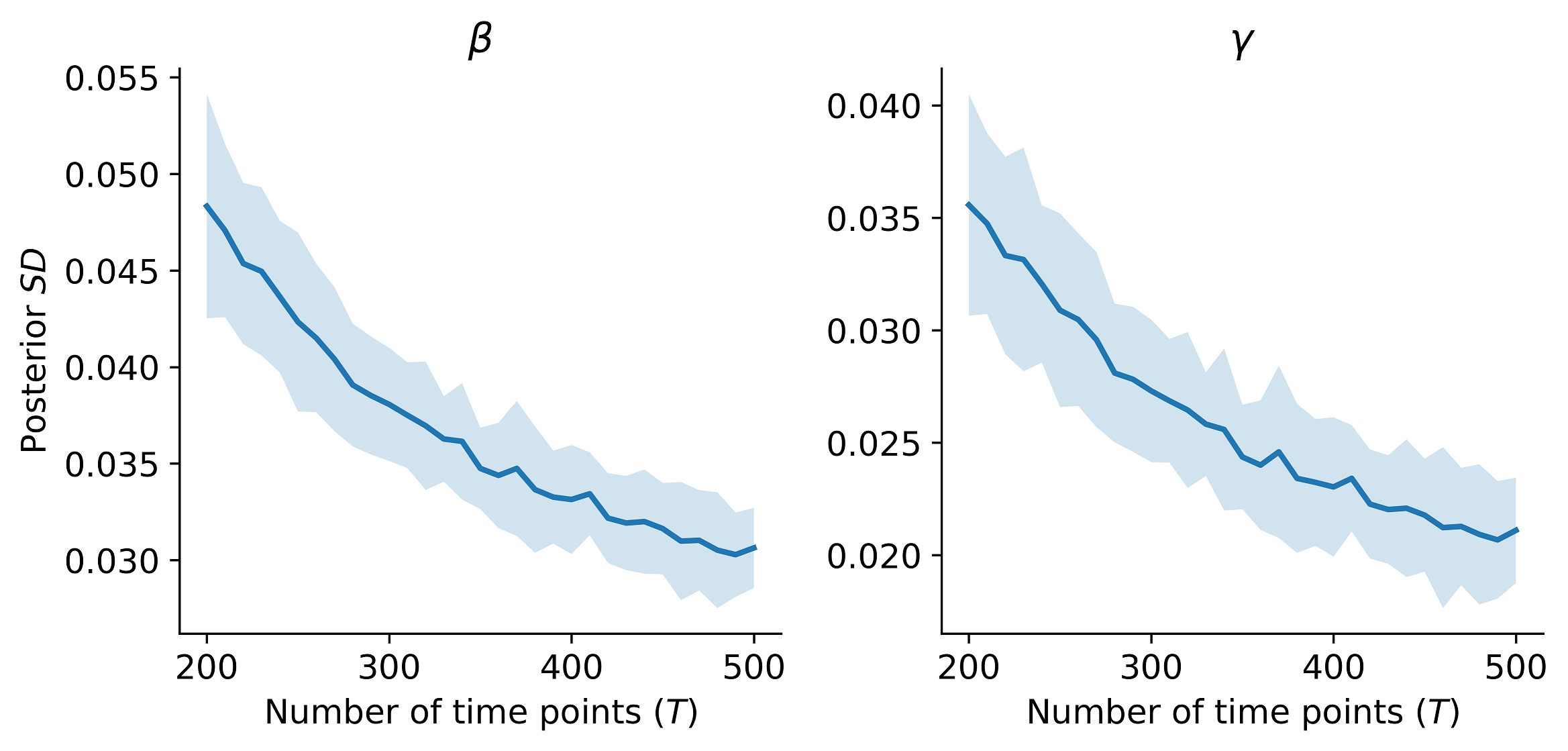}
    \caption{Posterior contraction over $T$}
    \label{fig:Fig.7d}
\end{subfigure}
\caption[short]{BayesFlow results obtained on the stochastic SIR model.} \label{fig:Fig.7}
\end{figure*}

During training of the networks, we simulate time-series from the stochastic SIR model with varying lengths. The number of time points $T$ is drawn from a uniform distribution $T \sim \mathcal{U}(200, 500)$ at each training iteration. For small $T$, the system has not yet reached an equilibrium (i.e., not all individuals have transitioned from being $I$ to $R$). It is especially interesting to see if BayesFlow can recover the rate parameters, while the process dynamics are still unfolding over time. Training the networks took approximately two hours with the online learning approach. Inference on $1000$ datasets with $2000$ posterior samples per parameter took approximately $1.1$ seconds. 

The results on the SIR model are depicted in \autoref{fig:Fig.7}. In line with the previous examples, we observe very good recovery of the true parameters, with NRMSE at $T=500$ around $0.03$, and $R^{2}$s around $0.99$. We observe decent performance even at smaller $T$s and the expected improvements as $T$ increases. Specifically, the posterior variance shrinks as $T$ increases, The SCB plots indicate that the approximate posteriors are well calibrated, with the approximate posterior mean of $\beta$ slightly overestimating the true parameter values in the lower range.

\subsection{Learned vs. Hand-Crafted Summaries: The Lotka-Volterra Population Model}

See \textbf{Appendix A}

\section{Discussion}

In the current work, we proposed and explored a novel method which uses invertible neural networks to perform globally amortized approximate Bayesian inference. The method, which we named BayesFlow, requires only simulations from a forward model to learn an efficient probabilistic mapping between data and parameters. We demonstrated the utility of BayesFlow by applying it to models and data from various research domains. Further, we explored an online learning approach with variable number of observations per iteration. We demonstrated that this approach leads to excellent parameter estimation throughout the examples considered in the current work. In theory, BayesFlow is applicable to any mathematical forward model which can be implemented as a computer simulation. In the following, we highlight the main advantages of BayesFlow. 

First, the introduction of separate summary and inference networks renders the method independent of the shape or the size of the observed data. The summary network learns a fixed-size vector representation of the data in an automatic, data-driven manner. Since the summary network is optimized jointly with the inference network, the learned data representation is encouraged to be maximally informative for inferring the parameters' posterior. This is particularly useful in settings where appropriate summary statistics are not known and, as a consequence, relevant information is lost through the choice of sub-optimal summary functions. However, if \textit{sufficient} statistics are available in a given domain, one might omit the summary network altogether and feed these statistics directly to the invertible network.

Second, we showed that BayesFlow generates samples from the correct posterior under perfect convergence without distributional assumptions on the shape of the posterior. This is in contrast to variational methods which optimize a lower-bound on the posterior \cite{kingma2016improved, kingma2014auto}, and oftentimes assume Gaussian approximate posteriors. Additionally, we also showed throughout all examples that the posterior means generated by the BayesFlow method are mostly excellent estimates for the true values.
Beyond this, the fact that the BayesFlow method recovers the full posterior over parameters does not necessitate the usage of point estimates or summary statistics of the posterior. 
Further, we observe the desired posterior contraction (posterior variance decreases with increasing number of observations) and better recovery with increasing number of observations. 
These are indispensable properties of any Bayesian parameter estimation method, since they mirror the decrease in epistemic uncertainty and the simultaneous increase in information due to availability of more data. 

Third, the largest computational cost of BayesFlow is paid during the training phase. 
Once trained, the networks can efficiently compute the posterior for any observed dataset arising from the forward model.
This is similar to the recently introduced {\em prepaid method} \cite{mestdagh2019prepaid}.
However, this method memorizes a large database of pre-computed summary statistics for fast nearest-neighbor inference, 
whereas a BayesFlow's network weights define an abstract representation of the relationship between data and parameters over the whole space of hidden parameters.
Traditionally, abstract representations like this only existed for analytically invertible model families, whereas more complex forward models required case-based inference, that is, expensive re-training for each observed dataset.
Amortized inference as realized by BayesFlow is thus especially advantageous for exploring, testing and comparing competing scientific hypotheses in research domains where an intractable model needs to be fit to multiple independent datasets.

Finally, all computations in the BayesFlow method benefit from a high degree of parallelism and can thus utilize the advantages of modern GPU acceleration.

These advantages notwithstanding, limitations of the proposed method should also be mentioned. Although we could provide a theoretical guarantee that BayesFlow samples from the true joint posterior under perfect convergence, this might not be achieved in practice. Therefore, is it essential that proper calibration of point estimates and estimated joint posteriors is performed for each application of the method. Fortunately, validating a trained BayesFlow architecture is easy due to amortized inference. Below, we discuss potential challenges and limitations of the method.

First, the design of the summary network and inference networks is a crucial choice for achieving optimal performance of the method. 
As already mentioned, the summary network should be able to represent the observed data without losing essential information and the invertible network should be powerful enough to capture the behavior of the forward model
Nevertheless, in some real-world scenarios, there might be little guidance on how to actually construct suitable summary networks. Recent work on probabilistic symmetry \cite{bloem2019probabilistic} and algorithmic alignment \cite{xu2019what} as well as our current experiments do, however, provide some insights about the design of summary networks. For instance, \textit{i.i.d.} data induce a permutation invariant distribution which is well modeled with a deep invariant network \cite{bloem2019probabilistic}. Data with temporal or spatial dependencies are best modeled with recurrent \cite{hwang2018conditional}, or convolutional \cite{radev2019towards} networks. When pairwise or multi-way relationships are particularly informative, attention \cite{vaswani2017attention} or graph networks \cite{xu2019what} appear as reasonable choices. On the other hand, the depth of the invertible network should be tailored to the complexity of the mathematical model of interest. More ACBs will enable the network to encode more complex distributions but will increase training time. Very high-dimensional problems might also require very large networks with millions of parameters, up to a point where estimation becomes practically unfeasible. However, most mathematical models in the life sciences prioritize parsimony and interpretability, so they do not contain hundreds or thousands of latent parameters. In any case, future applications might require novel network architectures and solutions which go beyond our initial recommendations. 

Another potential issue is the large number of neural network and optimization hyperparameters that might require fine-tuning by the user for optimal performance on a given task. We observe that excellent performance is often achieved with default settings. Using larger networks consisting of 5 to 10 ACBs does not seem to hurt performance or destabilize training, even if the model to be learned is relatively simple. Based on our results, we expect that a single architecture should be able to perform well on models from a given domain. Future research should investigate this question of generality by applying the method to different or even competing models within different research domains. Future research should investigate the impact of modern hyperparameter optimization methods such as Bayesian optimization \cite{eggensperger2013towards}.

Finally, even though modern deep learning libraries allow for rapid and relatively straightforward development of various neural network architectures, the implementational burden associated with the method is non-trivial. Thus, we are currently developing a general user-friendly software, which will abstract away most intricacies from the users of our method.

We hope that the new BayesFlow method will enable researchers from a variety of fields to accelerate model-based inference and will further prove its utility beyond the examples considered in this paper.


%

\section*{Acknowledgment}

We thank Paul Bürkner, Manuel Haussmann, Jeffrey Rouder, Raphael Hartmann, David Izydorczyk, Hannes Wendler, Chris Wendler, and Karin Prillinger for their invaluable comments and suggestions that greatly improved the manuscript. We also thank Francis Tuerlinckx and Stijn Verdonck for their support and thought-provoking ideas.



%
\bibliography{references}{}
\bibliographystyle{plain}

\clearpage

\section*{Appendix}
\setcounter{equation}{0}
\setcounter{figure}{0}
\setcounter{table}{0}
\setcounter{page}{1}
\setcounter{section}{1}
\makeatletter

\renewcommand{\thefigure}{S\arabic{figure}}
\renewcommand{\thesubsection}{\Alph{subsection}}

\subsection{Learned vs. Hand-Crafted Summaries: The Lotka-Volterra Population Model}

With this final example, we want to compare the performance of our method with an LSTM summary network vs. performance obtained with a standard set of hand-crafted summary statistics. For this, we focus on the well-studied Lotka-Volterra (LV) model. The LV model describes the dynamics of biological systems in which a population of predators interacts with a population of prey \cite{wilkinson2011stochastic}.  It involves a pair of first order, non-linear, differential equations given by:
\begin{align}
\frac{d}{dt} &= \alpha u - \beta u v \\
\frac{d}{dt} &=  -\gamma v + \delta \beta u v 
\end{align}
where $u$ denotes the number of preys, $v$ denotes the number of predators, and the parameter vector controlling the interaction between the species is $\bs{\theta} = (\alpha, \beta, \gamma, \delta)$. 

During training of the networks, we set the initial conditions as $u_0 = 10$ and $v_0 = 5$ and consider an interval $I_{T} = 15$ of discrete time units with $T=500$ time steps (samples) in between. Each sample $\bs{x}_t$ in each LV time-series $\bs{x}_{1:T}$ is thus a 2-dimensional vector containing the number of prey and predators in the population at time unit $t$.

We train two invertible neural networks. The first is trained jointly with an LSTM summary network which outputs a 9-dimensional learned summary statistic $h_{\bs{\psi}}(\bs{x}_{1:T})$. The second uses a set of 9 typically used, hand-crafted summary statistics \cite{papamakarios2016fast, papamakarios2018sequential}, which include: the mean of the time series; the log variance of the time-series; the auto-correlation of each timeseries at lags 0.2 and 0.4 time units; the cross-correlation between the two time series. The same cINN architecture with 5 ACBs is used for both training scenarios. For each scenario, we perform the same number of iterations and epochs. Online learning for each training scenario took approximately 4 hours in total wall-clock time.

\begin{figure}
\centering
\begin{subfigure}{.48\textwidth}
    \includegraphics[width=\textwidth]{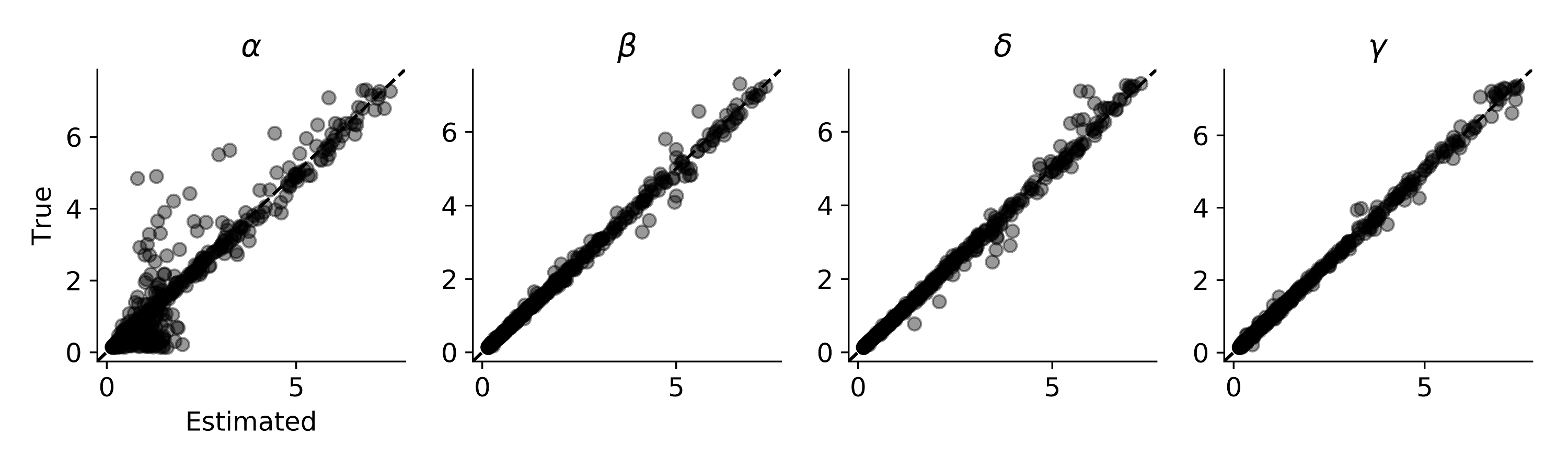}
    \caption{Parameter recovery with learned summary statistics}
    \label{fig:Fig.8a}
\end{subfigure}
\begin{subfigure}{.48\textwidth}
    \includegraphics[width=\textwidth]{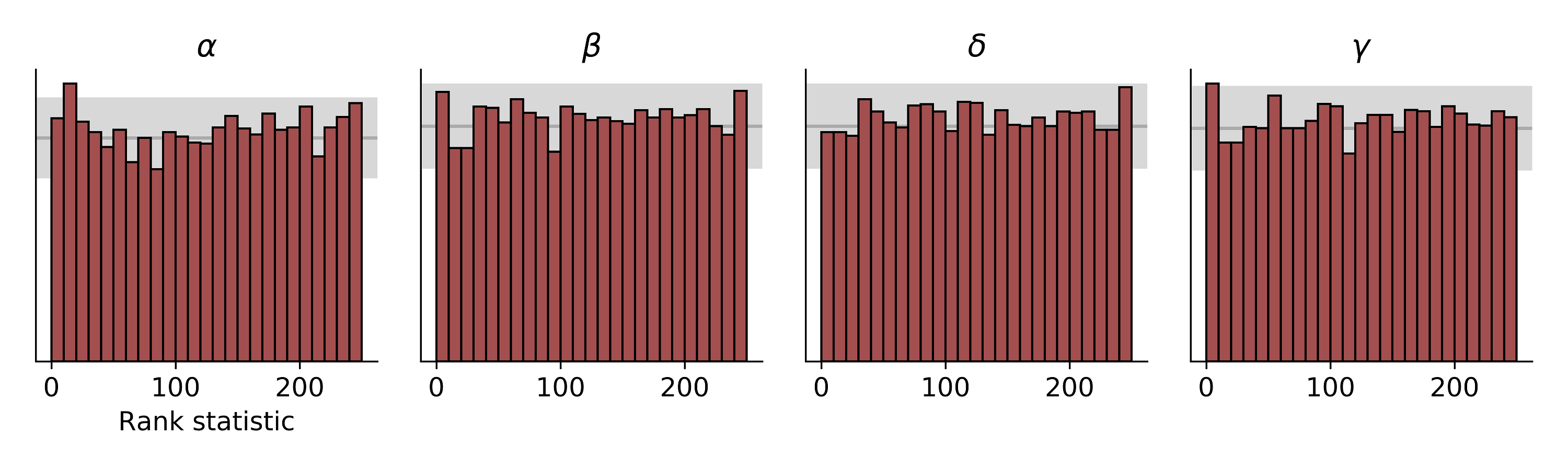}
    \caption{Calibration with learned summary statistics}
    \label{fig:Fig.8b}
\end{subfigure}
\begin{subfigure}{.48\textwidth}
    \includegraphics[width=\textwidth]{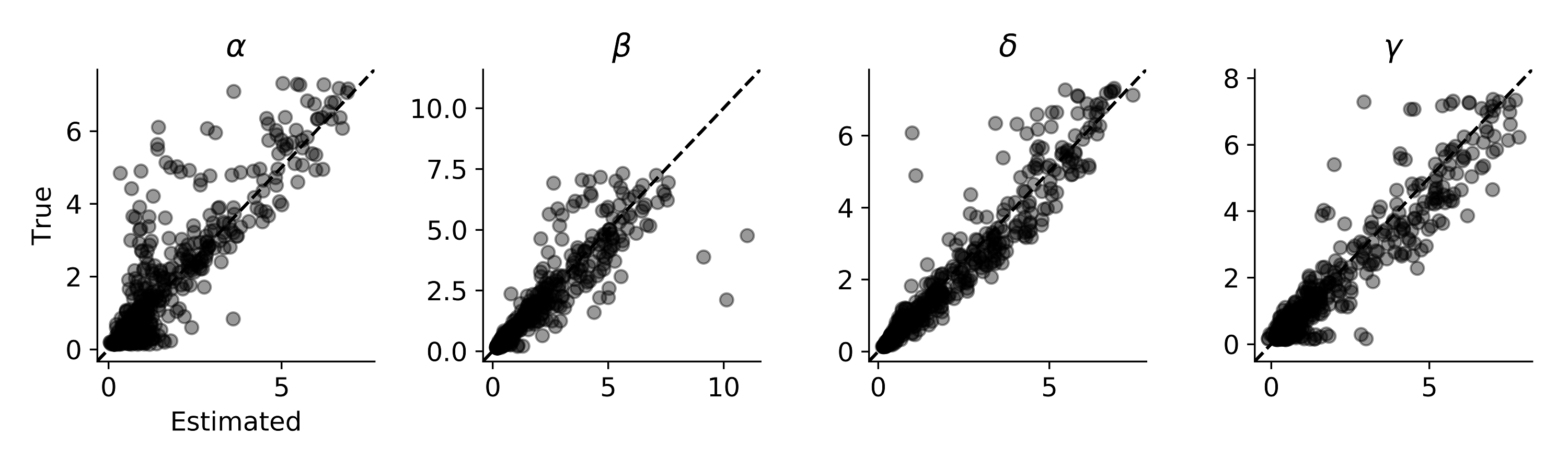}
    \caption{Parameter recovery with hand-crafted summary statistics}
    \label{fig:Fig.8c}
\end{subfigure}
\begin{subfigure}{.48\textwidth}
    \includegraphics[width=\textwidth]{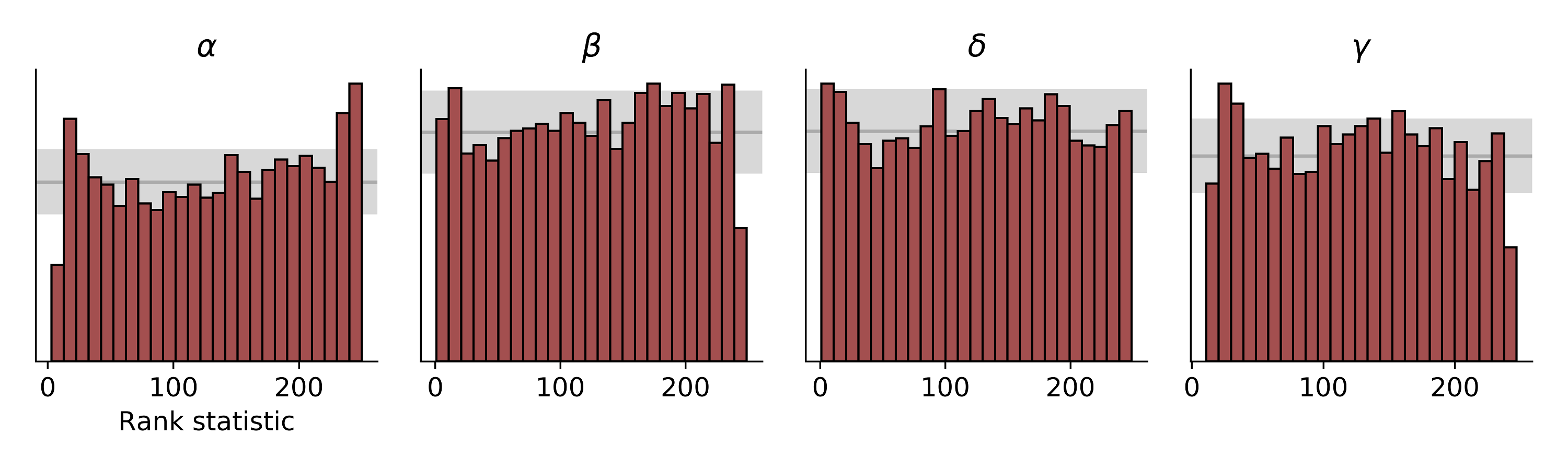}
    \caption{Calibration with hand-crafted summary statistics}
    \label{fig:Fig.8d}
\end{subfigure}
\begin{subfigure}{.48\textwidth}
    \includegraphics[width=\textwidth]{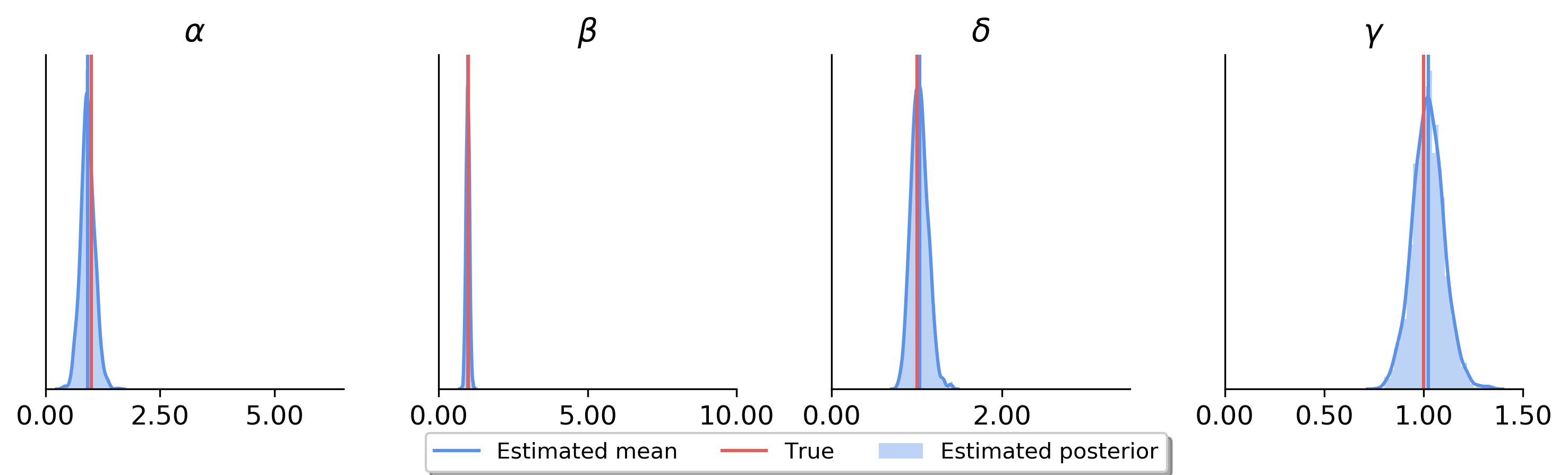}
    \caption{Full posterior with learned summary statistics}
    \label{fig:Fig.8e}
\end{subfigure}
\begin{subfigure}{.48\textwidth}
    \includegraphics[width=\textwidth]{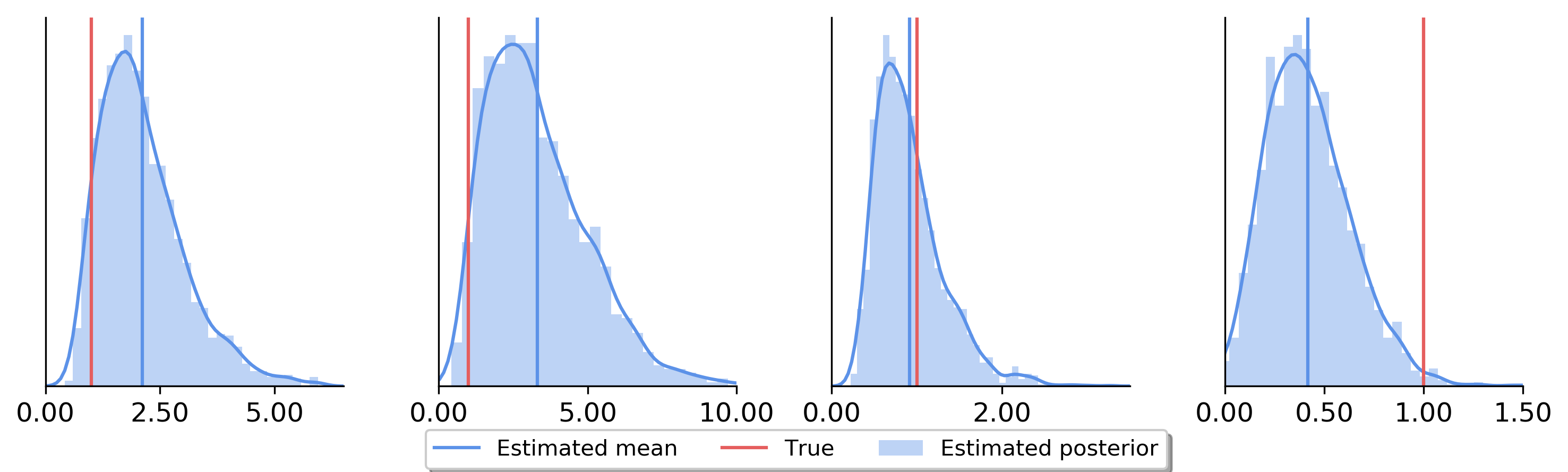}
    \caption{Full posterior with hand-crafted summary statistics}
    \label{fig:Fig.8f}
\end{subfigure}
\caption[short]{Comparison of recovery/calibration on the LV model with learned vs. hand-crafted summary statistics \textbf{(a)} Simulation-based calibration (SBC) with learned summary statistics; \textbf{(b)} Parameter recovery with learned summary statistics; \textbf{(c)} Parameter recovery with hand-crafted summary statistics; \textbf{(d)} Simulation-based calibration (SBC) with hand-crafted summary statistics; \textbf{(e)} Example full posteriors obtained on a single dataset with ground-truth rate parameters $\bs{\theta} = (1, 1, 1, 1)$ obtained with learned summaries; \textbf{(f)} The posterior obtained from the same dataset using hand-crafted summary statistics.} \label{fig:Fig.8}
\end{figure}

The results obtained on the LV model are depicted in \autoref{fig:Fig.8}. We observe notably better recovery of the true parameter estimates when performing inference with the learned summary statistics. The approximate posteriors are also better calibrated when conditioned on the set of 9 learned summary statistics. These results highlight the advantages of using a summary networks when no sufficient summary statistics are available. Finally, \autoref{fig:Fig.8e} and \autoref{fig:Fig.8f} depict the posteriors obtained by the two different INNs on a single dataset with ground-truth parameters $\bs{\theta} = (1,1,1,1)$. Evidently, learning the summary statistics leads to much sharper posteriors and better point-estimate recovery.

\subsection{Computation of Validation Metrics}

\subsubsection*{Normalized Root Mean Squared Error}
The normalized root mean squared error (NRMSE) between a sample of true parameters $\{\theta^{(m)}\}_{m=1}^M$ and a sample of estimated parameters $\{\hat{\theta}^{(m)}\}_{m=1}^M$ is given by:

\begin{align}
NRMSE = \sqrt{\sum_{m=1}^M\frac{\left(\theta^{(m)}-\hat{\theta}^{(m)}\right)^{2}}{\theta_{max}- \theta_{min}}} 
\end{align}

Due to the normalization factor $\theta_{max}-\theta_{min}$, the NRMSE is scale-independent, and thus suitable for comparing the recovery across parameters with different numerical ranges. The NRMSE equals zero when the estimates are exactly equal to the true values.

\subsubsection*{Coefficient of Determination }
The coefficient of determination $R^{2}$ measures the proportion of variance in a sample of true parameters $\{\theta^{(m)}\}_{m=1}^M$ that is \textit{explained} by a sample of estimated parameters $\{\hat{\theta}^{(m)}\}_{m=1}^M$. It is computed as:

\begin{align}
R^{2} = 1 - \sum_{m=1}^M\frac{\left(\theta^{(m)}-\hat{\theta}^{(m)}\right)^{2}}{\left(\theta^{(m)}-\bar{\theta}^{(m)}\right)^{2}} 
\end{align}
where $\bar{\theta}$ denotes the mean of the true parameter samples. When $R^{2}$ equals $1$, the estimates are perfect reconstructions of the true parameters.

\subsubsection*{Re-simulation Error}
To compute the re-simulation error $Err_{sim}$, we first obtain an estimate of the true parameter value given an observed (validations) dataset $\bs{x}^{o}_{1:N}$ by computing the mean of the approximate posterior $\tilde{\bs{\theta}}$. Then, we run the mathematical model to obtain a simulated dataset $\bs{x}^{s}_{1:N} = g(\tilde{\bs{\theta}},\xib)$. Finally, we compute the maximum mean discrepancy (MMD, \cite{gretton2012kernel}) between the observed and the simulated dataset $MMD(\bs{x}^{o}_{1:N},\bs{x}^{s}_{1:N})$. The MMD is a kernel-based metric which estimates the mismatch between two distributions given samples from the distributions by comparing all of their moments. It equals zero when the two distributions are equal almost everywhere \cite{gretton2012kernel}). Thus, a low MMD indicates that the distribution of $\bs{x}_{1:N}^{s}$ is close to the distribution of $\bs{x}^{o}_{1:N}$. Conversely, a high MMD indicates that the distribution of $\bs{x}^{s}_{1:N}$ is far from the distribution of $\bs{x}^{o}_{1:N}$. We report the median MMD computed over all validation datasets.   

\subsubsection*{Calibration Error}
The calibration error $Err_{cal}$ quantifies how well the coverage of an approximate posterior matches the coverage of an unknown true posterior. Let $\alpha_{\theta}$ be the fraction of true parameter values lying in a corresponding $\alpha$-credible interval of the approximate posterior. Thus, for a perfectly calibrated approximate posterior, $\alpha_{\theta}$ should equal $\alpha$ for all $\alpha \in (0,1)$. We compute the calibration error for each marginal posterior as the median absolute deviation $\,|\,\alpha_{\theta}-\alpha\,|\,$ for 100 equally spaced values of $\alpha \in (0,1)$. Therefore, the calibration error ranges between 0 and 1 with 0 indicating perfect calibration and 1 indicating complete miscalibration of the approximate posterior.

\subsubsection*{Kullback-Leibler Divergence} 

The Kullback-Leibler divergence ($\mathbb{KL}$) quantifies the increase in entropy incurred by approximating a target probability distribution $P$ with a distribution $Q$. Its general form for absolutely continuous distributions is given by
\begin{align}
\mathbb{KL}(P\,||\,Q) = \int_{-\infty}^{\infty} p(x)\log\frac{p(x)}{q(x)} dx 
\end{align}

where $p$ and $q$ denote the pdfs of $P$ and $Q$. In the case where $P$ and $Q$ are both multivariate Gaussian distributions, the KL divergence can be computed in closed form \cite{hershey2007approximating}:

\begin{equation}
\begin{split}
\mathbb{KL}(P\,||\,Q) = \frac{1}{2}\left[\log\frac{\det\bs{\Sigma}_{q}}{\det\bs{\Sigma}_{p}} + \Tr(\boldsymbol{\Sigma}_{q}^{-1}\boldsymbol{\Sigma}_{p}) - d + (\boldsymbol{\mu}_{p} - \right. \\ \left. \bs{\mu}_{q})^{T}\boldsymbol{\Sigma}_{q}^{-1}(\bs{\mu}_{p} - \bs{\mu}_{q})\right] \label{eq:kl}
\end{split}
\end{equation}

where $\boldsymbol{\Sigma}_{p}$ and $\boldsymbol{\Sigma}_{q}$ denote the covariance matrices of $p$ and $q$, $\boldsymbol{\mu}_{p}$ and $\boldsymbol{\mu}_{q}$ the respective mean vectors, and $d$ the number of dimensions of the Gaussian. In the case of diagonal Gaussian distributions, Eq.\ref{eq:kl} reduces to:

\begin{align}
\mathbb{KL}(P\,||\,Q) = \sum_{i=1}^d\left(\log\frac{\sigma_{q,i}}{\sigma_{p,i}} + \frac{\sigma_{p,i}^{2} + (\mu_{q,i} - \mu_{p,i})^{2}}{2\sigma_{q,i}^{2}} - \frac{1}{2} \right) 
\end{align}

Even though the KL divergence is not a proper distance metric, as it is not symmetric in its arguments, it can be used to quantify the error of approximation when a closed-form solution is available.

\subsection*{Simulation-Based Calibration}
Simulation-based calibration is a method to detect systematic biases in any Bayesian posterior sampling method \cite{talts2018validating}. It is based on the \textit{self-consistency} of the Bayesian joint distribution. Given a sample from the prior distribution $\tilde{\theta} \sim p(\theta)$ and a sample from the forward model
$\tilde{x} \sim p(x\,|\,\tilde{\theta})$, one can integrate $\tilde{\theta}$ and $\tilde{x}$ out of the joint distribution and recover back the prior of $\theta$:

\begin{align}
p(\theta) &= \int p(\theta,\tilde{\theta},\tilde{x})d\tilde{x}d\tilde{\theta} \\
&= \int p(\theta,\tilde{x}\,|\,\tilde{\theta})p(\tilde{\theta})d\tilde{x}d\tilde{\theta} \\
&= \int p(\theta\,|\,\tilde{x})p(\tilde{x}\,|\,\tilde{\theta})p(\tilde{\theta})d\tilde{x}d\tilde{\theta} \label{eq:sbc}
\end{align}

If the Bayesian sampling method produces samples from the exact posterior, the equality implied by Eq.\ref{eq:sbc} should hold regardless of the particular form of the posterior. Thus, any violation of this equality indicates some error incurred by the sampling method. The authors of \cite{talts2018validating} propose \textbf{Algorithm} \ref{alg:2} for visually detecting such violations:

\begin{algorithm}
\caption{Simulation-based calibration (SBC) for a single parameter $\theta$}\label{alg:2}
\begin{algorithmic}[1]
\For{$m = 1,...,M$}
\State {Sample $\tilde{\theta}^{(m)} \sim p(\theta)$}
\State {Simulate a dataset $\boldsymbol{x}_{1:N}^{(m)} = g(\tilde{\theta}^{(m)},\xib)$}
\State {Draw posterior samples $\{\theta^{(l)}\}_{l=1}^{L} \sim p_{\boldsymbol{\phi}}(\theta\,|\,\boldsymbol{x}^{(m)}_{1:N})$} 
\State {Compute rank statistic $r^{(m)} = \sum_{l=1}^{L}\mathbbm{1}_{[\theta^{(l)} < \tilde{\theta}^{(m)}]}$}
\State {Store $r^{(m)}$}
\EndFor
\State {Create a histogram of $\{r^{(i)}\}_{m=1}^M$ and inspect it for uniformity}
\end{algorithmic}
\end{algorithm} 

\textbf{Algorithm} \ref{alg:2} is correct, since Eq.\ref{eq:sbc} implies that the rank statistic defined in line $5$ should be uniformly distributed. Hence, any deviations from uniformity indicate some interpretable error in the approximate posterior \cite{talts2018validating}.

\subsection{Model Details}

\subsubsection*{The Ricker Model}

\textit{Summary Network}. We use a bidirectional long short-term memory (LSTM) recurrent neural network  \cite{gers1999learning} for the raw Ricker time-series. The LSTM network architecture is a reasonable choice for this example, as it is able to capture long-term dependencies in datasets with temporal or spatial autocorrelations. LSTMs can also easily deal with variable-length time-series.

\textit{Simulation}. We place the following uniform priors over the Ricker model parameters:

\begin{align}
\rho &\sim \mathcal{U}(0,15) \\
r &\sim \mathcal{U}(1,90)   \\
\sigma &\sim \mathcal{U}(0.05,0.7)  
\end{align}

These ranges appear to be very broad, as datasets generated by extreme parameter values appear implausible in real-world scenarios. Nevertheless, we stick to broad priors for training, even though parameter recovery might degrade at the extremes.

\begin{figure}[H]
\centering
\begin{subfigure}{.99\textwidth}
	\includegraphics[width=\textwidth]{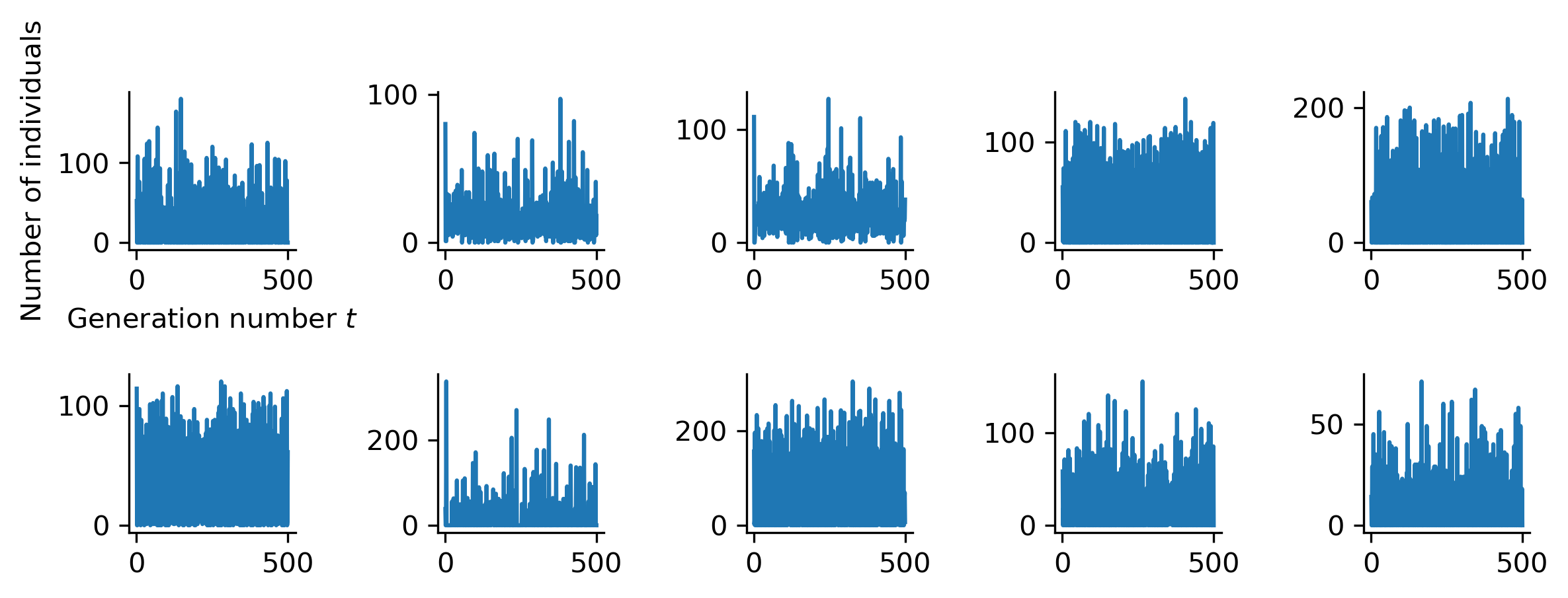}
\end{subfigure}
\caption{Example Ricker datasets generated with different parameters.} 
\label{fig:Fig.S2}
\end{figure}

\autoref{fig:Fig.S2} depicts different simulated Ricker timeseries generated via draws from the prior.

\subsubsection*{The Lévy-Flight Model}

\textit{Summary Network}. We use a permutation invariant neural network \cite{bloem2019probabilistic} for the \textit{i.i.d.} reaction times (RT) data. Similarly to the toy Regression example, each response in an RT dataset is assumed to be independent of all others, so permutations of the dataset must lead to the same parameter estimates. 

\textit{Simulation}. We place the following uniform priors over the LFM parameters, since they are broad enough to cover the range of realistic RT distributions encountered in empirical choice RT scenarios:

\begin{align}
v_{0} &\sim \mathcal{U}(0, 6)  \\
v_{1} &\sim \mathcal{U}(-6, 0)  \\
zr &\sim \mathcal{U}(0.3, 0.7)  \\
a &\sim \mathcal{U}(0.6, 3) \\
t_{0} &\sim \mathcal{U}(0.3, 1)  \\
\alpha &\sim \mathcal{U}(1, 2)
\end{align}

\begin{figure}[H]
\centering
\begin{subfigure}{.99\textwidth}
	\includegraphics[width=\textwidth]{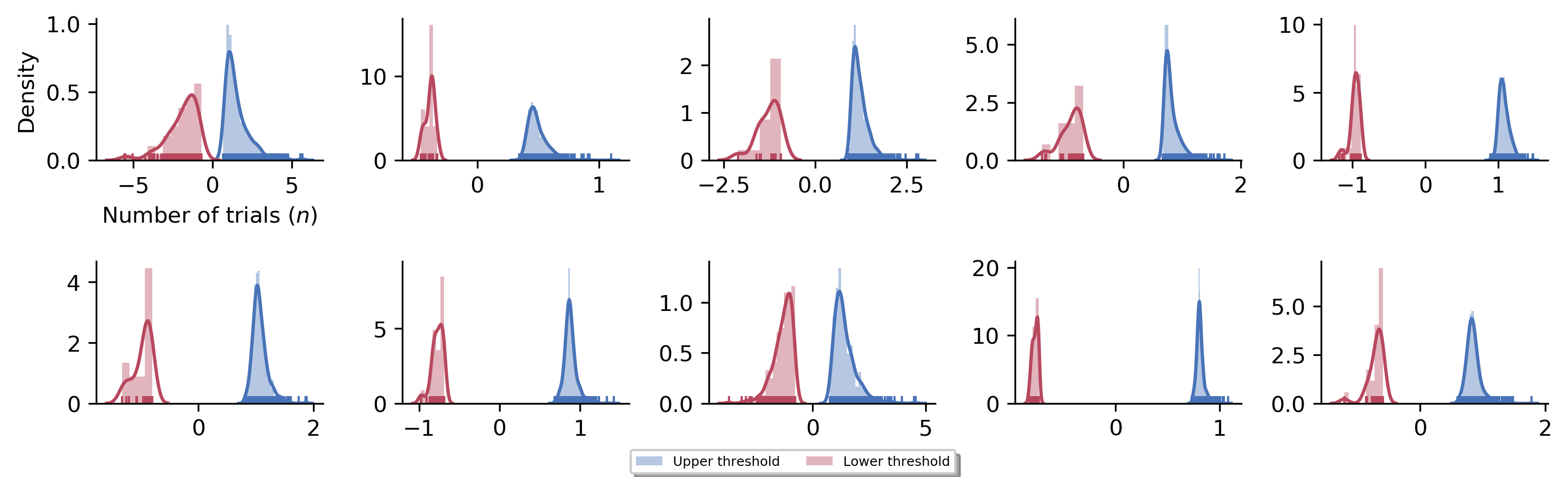}
\end{subfigure}
\caption{Example RT distributions generated with different parameters.} 
\label{fig:Fig.S3}
\end{figure}

\autoref{fig:Fig.S3} depitcs different simulated RT distributions generated via draws from the prior.

\subsubsection*{The Stochastic SIR Model}

\textit{Summary Network}. We use a $1D$ fully convolutional neural network \cite{long2015fully} for the raw SIR time-series into fixed-size vectors. Here, we choose a convolutional network architecture over the previously mentioned LSTM, as convolutional networks are more computationally efficient. Further, we wanted to underline the utility of 1D convolutional networks for multidimensional time-series data. Finally, convolutional networks can also deal with variable input sizes. 

\textit{Simulation}. We place the following uniform priors over the two rate parameters of the stochastic SIR model:

\begin{align}
\beta &\sim \mathcal{U}(0.01, 1) \\
\gamma &\sim \mathcal{U}(0.01,\beta) 
\end{align}

These ranges were chosen based on empirical plausibility of the generated SIR time-series. 

\begin{figure}[H]
\centering
\begin{subfigure}{.99\textwidth}
	\includegraphics[width=\textwidth]{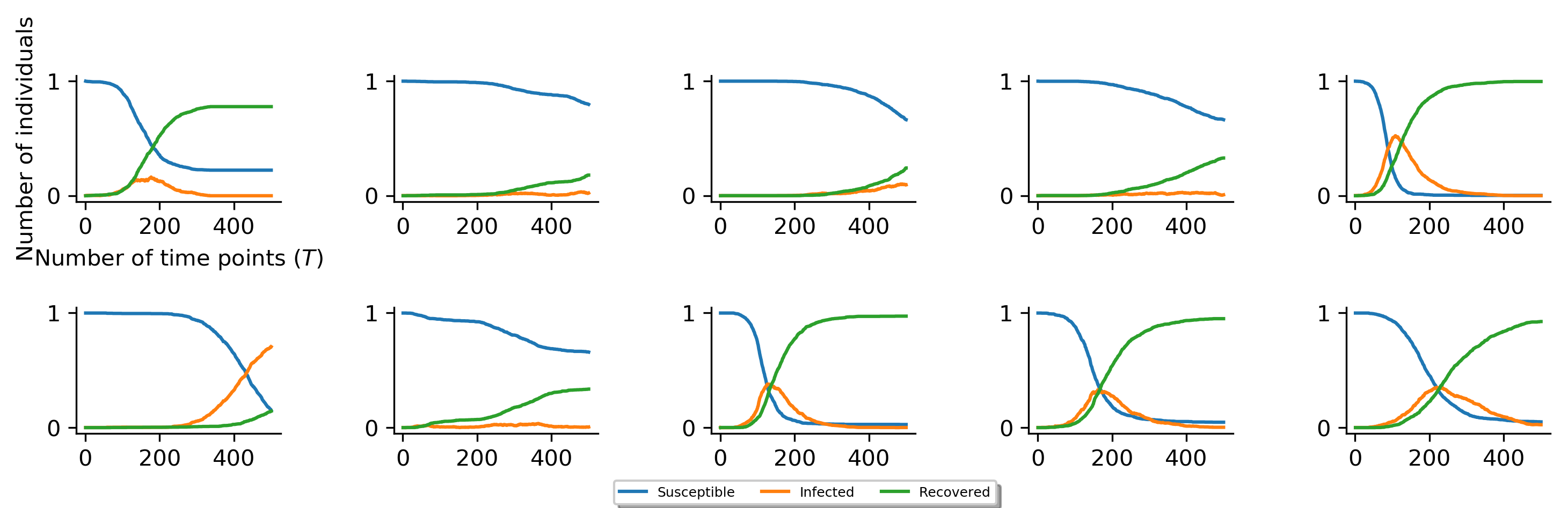}
\end{subfigure}
\caption{Example SIR timeseries generated with different parameters.} 
\label{fig:Fig.S4}
\end{figure}

\autoref{fig:Fig.S4} depicts different SIR timeseries generated via draws from the prior.

\subsubsection*{The Lotka-Volterra Model}

\textit{Summary Network}. We use a bidirectional long short-term memory (LSTM) recurrent neural network  \cite{gers1999learning} for the raw LV time-series (as in the Ricker example).

\textit{Simulation}. We place the following broad uniform priors over the LV parameters. Some of the parameter combinations produced divergent simulations, which we removed during online learning.

\begin{align}
\alpha &\sim \mathcal{U}(\exp{(-2)}, \exp{(2)})  \\
\beta &\sim \mathcal{U}(\exp{(-2)}, \exp{(2)}) \\
\gamma &\sim \mathcal{U}(\exp{(-2)}, \exp{(2)})  \\
\delta &\sim \mathcal{U}(\exp{(-2)}, \exp{(2)}) 
\end{align}

\subsection{Example Posteriors on Ricker Datasets}

\begin{figure}[H]
\centering
\begin{subfigure}{.99\textwidth}
	\includegraphics[width=\textwidth]{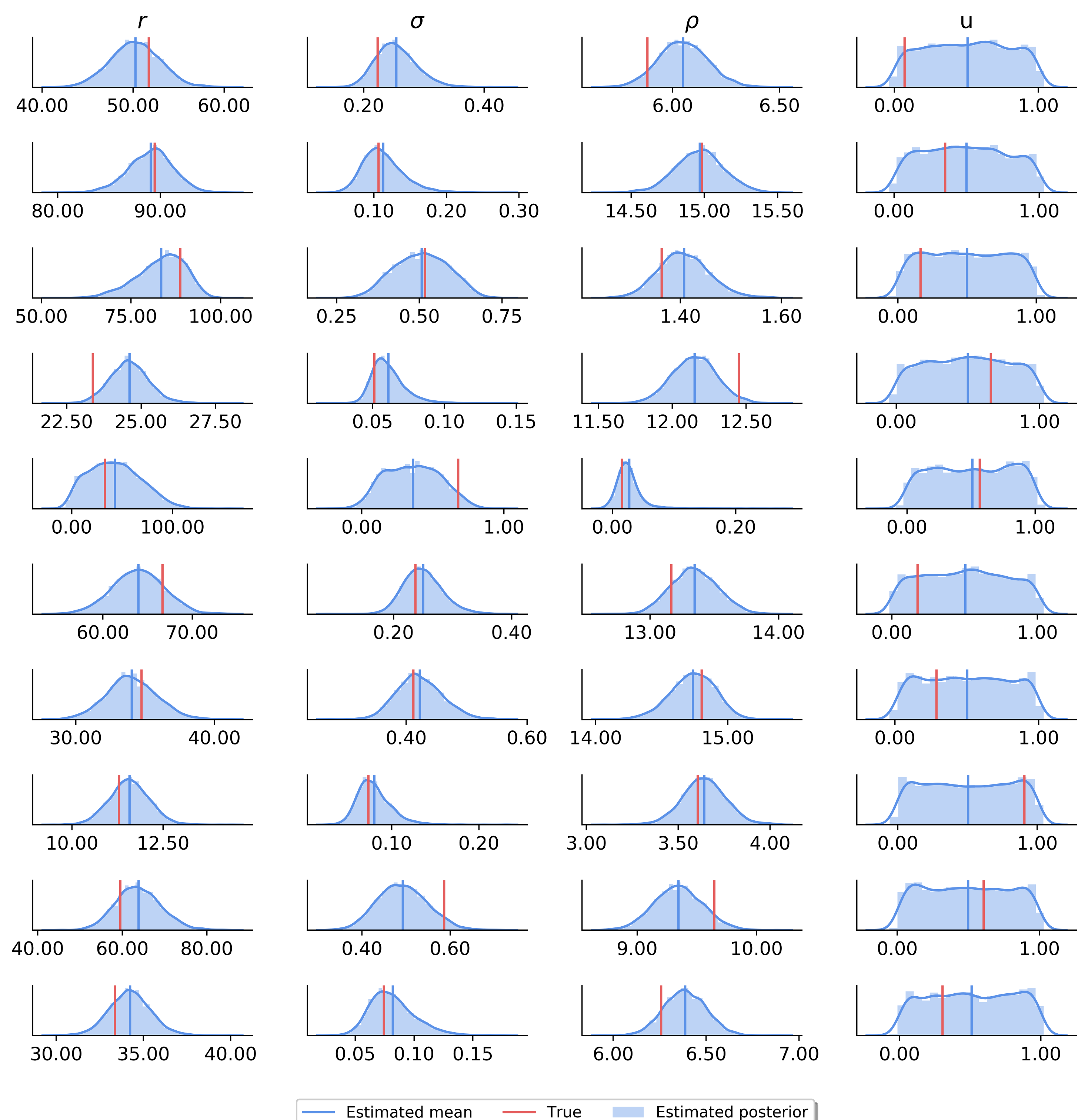}
\end{subfigure}
\caption{Ten example Ricker marginal posteriors} 
\label{fig:Fig.S6}
\end{figure}

Marginal posteriors from ten validation datasets simulated from the Ricker model are depicted in \autoref{fig:Fig.S6}. We observe widely different posterior shapes, highlighting the importance of working with arbitrary posterior shapes.

\end{document}